\documentclass[lettersize,journal]{IEEEtran}
\usepackage{times}

\usepackage{multirow}
\usepackage{amsmath} 
\usepackage{multicol}
\usepackage{graphicx}
\usepackage{xspace}
\usepackage{xcolor}
\usepackage{caption}
\usepackage{wrapfig}
\usepackage{bbding}  
\usepackage{pifont}  
\usepackage{bbm} 
\usepackage{hyperref}
\usepackage[capitalise, nameinlink]{cleveref}

\usepackage{booktabs}
\usepackage{arydshln} 
\usepackage{siunitx} 
\usepackage{float}
\usepackage{amssymb}  
\usepackage{bm}
\usepackage[table]{xcolor}
\usepackage{dsfont}
\usepackage{siunitx}
\newcommand{\ours}[0]{\textsc{Hyper-GNC}\xspace}

\newcommand{\ci}[1]{%
    \textcolor{gray}{%
        \tiny~($\pm \num{#1}$)%
    }%
}
\usepackage{graphicx}

\usepackage[numbers]{natbib}

\newcommand{\ourrow}{\rowcolor{gray!7}}

\definecolor{citecolor}{HTML}{faa700} 
\definecolor{lblue}{HTML}{ffb114} 
\definecolor{ogreen}{HTML}{2E7D32}
\definecolor{bred}{HTML}{BF360C}
\definecolor{newbrown}{HTML}{795548}

\hypersetup{
    colorlinks=true,
    linkcolor=bred,
    filecolor=magenta,      
    urlcolor=bred,
    citecolor=bred,
}

\usepackage{xfrac}

\begin{document}

\title{Learning Adaptive Multi-Task Guidance, Navigation, and Control via Hypernetworks}

\author{\IEEEauthorblockN{Ricard M. Castan\textsuperscript{1} \quad Aman Arora\textsuperscript{1} \quad Antoine Richard\textsuperscript{1} \quad Andrej Orsula\textsuperscript{1} \quad Cédric Pradalier\textsuperscript{2} \quad Miguel A. Olivares-Méndez\textsuperscript{1}}\\
\IEEEauthorblockA{\textsuperscript{1}University of Luxembourg \quad \textsuperscript{2}Georgia Institute of Technology}
}

\maketitle

\begin{abstract}
Autonomous free-flying robots in orbital environments require controllers that are both versatile and resource-efficient, yet maintaining a separate, task-specific policy for each mission profile is architecturally brittle and limits operational flexibility as requirements evolve. We introduce HYPER-GNC, a multi-task reinforcement learning framework in which a hypernetwork maps physics-informed task embeddings to the weights of a shared actor-critic policy, enabling a single compact controller to master four distinct GNC tasks: velocity tracking, docking, inspection, and navigation with obstacle avoidance. The continuous embedding space allows the controller to generalize to novel mission configurations at deployment time without any retraining. Extensive experiments demonstrate that HYPER-GNC achieves sample efficiency comparable to single-task specialists while maintaining stability under significant inertial perturbations and external body wrenches. We further validate the framework on a physical satellite emulator, successfully bridging the simulation-to-reality gap across all mission profiles. Code, trained models, and deployment scripts are made publicly available to support reproducibility.

\end{abstract}


\IEEEpeerreviewmaketitle

\section{Introduction}

\IEEEPARstart{T}{he} next era of space exploration demands a paradigm shift in orbital autonomy. As we move toward complex, multi-agent environments like the International Space Station (ISS) and the future Lunar Gateway, mobility in space becomes a fundamental requirement. Free-flying robots, such as JAXA's Int-Ball2 \cite{Mitani2023IntBall2, yamaguchi2024intball2}, are uniquely suited for these environments, offering six-degree-of-freedom (6-DOF) movement to assist astronauts, conduct inspections, and manage logistics. However, controlling these free-flyers across a diverse mission set presents a significant bottleneck. Maintaining a library of discrete controllers for docking, tracking velocities, inspection, and navigation with obstacle avoidance is computationally expensive and architecturally brittle.

Beyond architectural complexity, this dependence on discrete, mission-specific controllers imposes a hard ceiling on operational mission lifetime. Each time mission requirements evolve---an unexpected obstacle field, an unplanned docking rendezvous, or a contingency inspection---ground operators must upload a new or reconfigured controller, incurring communication delays, mission downtime, and windows of reduced autonomy. A unified controller capable of synthesizing novel behavioral combinations from a compact, pre-learned repertoire would fundamentally change this paradigm: the robot could autonomously adapt to evolving mission profiles without ground intervention, directly extending its effective operational lifetime.

\begin{figure}[t]
    \centering
    \includegraphics[width=0.5\textwidth]{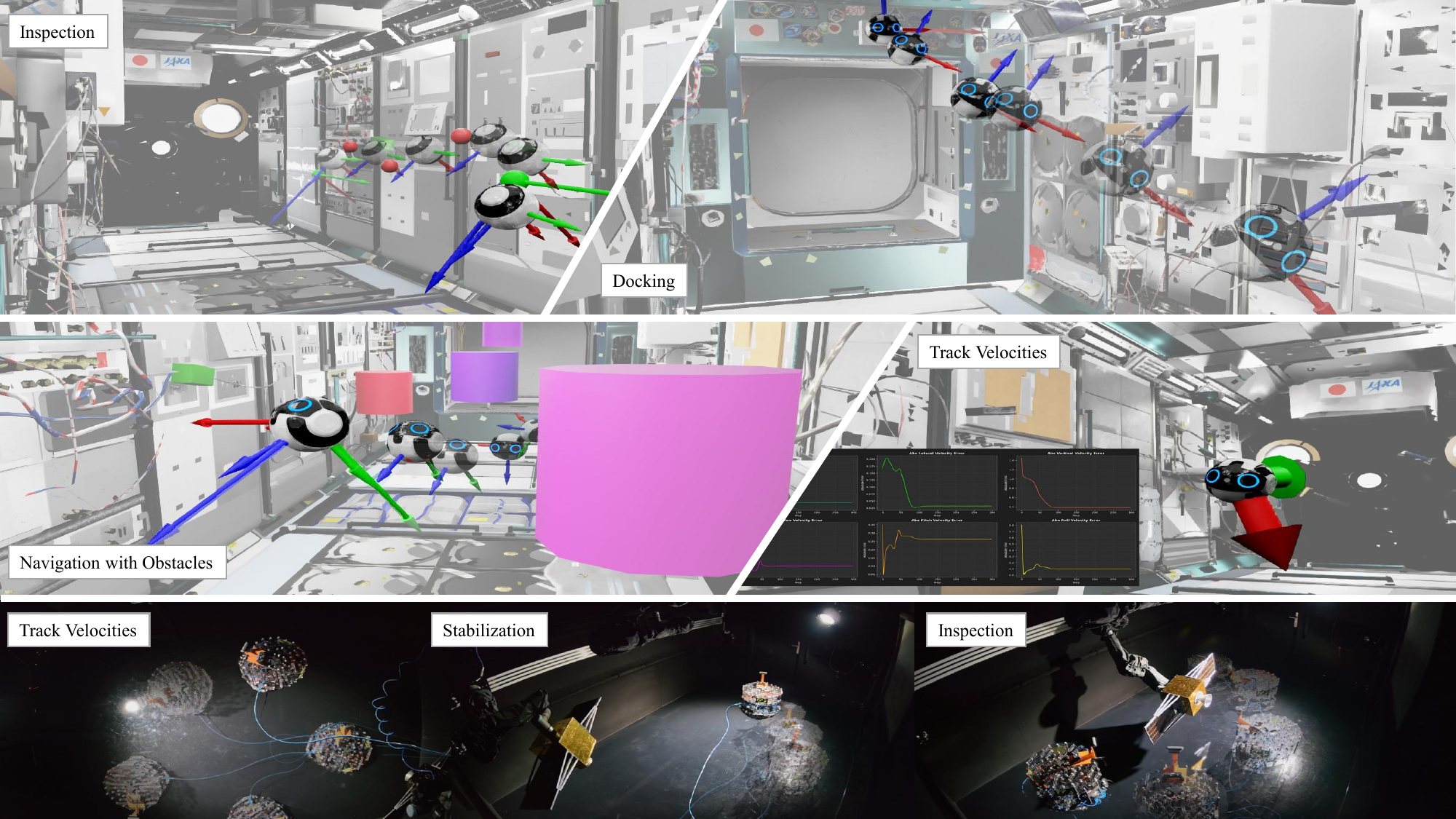}
    \vspace{-0.17in}
    \caption{\textbf{Overview.} Our proposed framework, \ours, performing distinct GNC tasks in simulated environments (top two rows) and transferring the framework for sim-to-real in a satellite emulator (bottom row).}
    \label{fig:teaser}
    \vspace{-1.5em}
\end{figure}

Reinforcement Learning (RL) offers a data-driven alternative capable of mastering complex, non-linear dynamics. Yet, standard on-policy RL is notoriously data-inefficient ~\cite{schulman2017trustregionpolicyoptimization, schulman2017proximalpolicyoptimizationalgorithms}, and often results in specialist models that fail to generalize to novel scenarios~\cite{ghosh2021generalizationrldifficultepistemic, cobbe2019quantifyinggeneralizationreinforcementlearning, pmlr-v97-cobbe19a}. To mitigate these limitations, Multi-Task Reinforcement Learning (MTRL) emerges as a promising solution~\cite{Caruana1997Multitask, ruder2017overviewmultitasklearningdeep, zhang2021surveymultitasklearning}. By focusing on learning a single, generalized policy capable of adapting to diverse tasks, MTRL aims to improve data efficiency and leverage shared knowledge across related tasks. This paradigm facilitates the transfer of learned behaviors and representations~\cite{parisotto2016actormimicdeepmultitasktransfer}, thereby enhancing the overall learning process and reducing the need for extensive, task-specific retraining ~\cite{andreas2017modularmultitaskreinforcementlearning, devin2016learningmodularneuralnetwork}.

The primary challenge in multi-task learning is designing a controller that can generalize across different scenarios, as this requires balancing conflicting objectives and diverse state spaces~\cite{yu2020gradientsurgerymultitasklearning, chen2018gradnormgradientnormalizationadaptive, beck2025tutorialmetareinforcementlearning}. For example, a robot's need for stabilization during docking maneuvers directly conflicts with the objective of increasing speed for obstacle avoidance. This contradiction underscores the difficulty of creating a single policy that can efficiently manage multiple and sometimes opposing goals.

 Furthermore, implementing MTRL presents its own architectural challenges. The straightforward solution of simply scaling up model capacity (a single, large network) or naïvely merging datasets can often lead to unstable training or severe conflicting gradients between tasks, ultimately degrading performance~\cite{yu2020gradientsurgerymultitasklearning}. Crucially, in our target domain of satellite control, deploying large, monolithic models is often infeasible. These environments demand not only a generalized policy but one that is also compact, memory-efficient, and computationally lean to ensure reliable and real-time operation under severe constraints \cite{9600851}. Therefore, an MTRL architecture is required that can achieve generalization and shared knowledge without resorting to excessive model size or sacrificing training stability.

We consider the problem of learning a unified Guidance, Navigation, and Control (GNC) policy for a free-flying spacecraft. Such a policy must master distinct and complex mission profiles while synthesizing novel combinations of behaviors on demand, thereby extending operational mission lifetimes by reducing dependence on ground-based controller updates. We introduce \ours, an MTRL framework for GNC policies. \ours\ uses physics-informed semantic embeddings to resolve potential task conflicts and enable robust behavioral composition for novel mission configurations without any on-orbit retraining.

While the ability to handle novel scenarios is often associated with Meta-Reinforcement Learning (Meta-RL), we intentionally adopt a MTRL framework driven by a Hypernetwork architecture. Standard Meta-RL typically relies on fast adaptation, requiring the agent to perform additional trial-and-error transitions or maintain complex recursive hidden states to infer a new task at test time \cite{finn2017modelagnosticmetalearningfastadaptation, yu2021metaworldbenchmarkevaluationmultitask}. Such processes introduce non-deterministic behavior and computational overhead that are often unacceptable for the strict safety protocols of orbital robotics. Instead, \ours\ achieves adaptation through semantic interpolation. By conditioning the Hypernetwork on physics-informed embeddings, our model treats the diverse mission profiles not as isolated silos, but as points within a continuous task manifold. This allows the controller to compose and recombine its learned behavioral repertoire to address novel mission configurations at deployment time, without requiring any on-line retraining or ground intervention. This approach provides the flexibility of Meta-RL with the architectural stability and immediate deployment readiness of a fixed MTRL policy, making it well-suited to sustaining autonomous operation across the full arc of an orbital mission lifetime.

We model in simulation JAXA's Int-Ball2, a spherical free-flying robot currently operating aboard the ISS. We train this model to execute the following real-world mission profiles: Docking, essential for station-keeping and autonomous battery charging; Velocity Command Tracking, required for high-precision teleoperation or cooperative manipulation tasks; Inspection, used for supporting scientific objectives by recording astronaut experiments from specific, controlled angles; and Navigation with Obstacle Avoidance, which is critical for safe, autonomous movement in the crowded ISS environment to prevent collisions with floating objects and critical equipment.

We also show that our framework can be used in a real-world physical satellite emulator or Floating Platform (FP) to bridge the reality gap. To ensure real-world deployability, we apply comprehensive Domain Randomization \cite{8202133} across key physical parameters, including inertial properties, and body wrenches.

Finally, we release our simulation MTRL framework\footnote{Code:\url{https://github.com/snt-spacer/Hyper-GNC.git}}, allowing other practitioners to plug in their robots easily and test their environments to further facilitate multi-task RL research. We also release our weights and our sim-to-real code for reproducibility \footnote{Weights:\url{https://huggingface.co/r3m3c3/HyperGNC/tree/main}} \footnote{Sim2Real:\url{https://github.com/snt-spacer/Hyper-GNC_Deployment.git}}.

\section{Related Work}

\subsection{Reinforcement Learning for Spacecraft GNC}

The application of Deep Reinforcement Learning (DRL) to spacecraft GNC has gained significant traction due to its ability to handle high-dimensional state spaces and model non-linear dynamics \cite{federici2025optical, izzo2019survey, Gaudet_2020}.
Prior work has largely focused on mastering specific, isolated mission phases.

In the domain of planetary landing, \citet{gaudet2020deep} demonstrated that DRL can achieve pinpoint 6-DoF landing accuracy with fuel optimality that surpasses convex optimization methods.
Similarly, for small-body operations, studies have successfully applied RL in simulation to autonomous maneuvering and mapping around asteroids \cite{willis2016reinforcement, chan2019autonomous}.
Regarding rendezvous and docking, recent approaches have utilized DRL for collision-free trajectory generation in cluttered environments \cite{athauda2023intelligent} and robust proximity operations under uncertainty \cite{hovell2021deep}.
Furthermore, specialized policies have been developed for attitude stabilization and slew maneuvers, offering faster settling times than classical controllers \cite{El_Hariry_2024}.

However, these approaches typically treat mission phases as separate control problems.
A policy trained for the high-speed dynamics of inspection often fails when applied to the precision constraints of docking.
While recent benchmarking efforts have begun to formalize the need for generalist robot learning in space \cite{orsula2025space}, and meta-learning approaches have been validated for visual adaptation on optical benches \cite{federici2025optical}, a unified GNC framework remains elusive.
The current standard, a disjointed ensemble of experts \cite{hendawy2023multi}, increases on-board memory requirements and prevents the transfer of shared physical dynamics, such as inertia tensor handling and actuator delays, across tasks.

\subsection{Multi-Task Reinforcement Learning in Robotics}

MTRL aims to learn a single generalist policy capable of executing diverse behaviors, theoretically improving sample efficiency through positive transfer \cite{teh2017distralrobustmultitaskreinforcement, kalashnikov2021scaling}.
The most straightforward approach is Goal-Conditioned RL, where a task identifier (TaskID) or target state is appended to the observation vector \cite{Kaelbling1993LearningTA}.
However, in heterogeneous robotic domains, this approach suffers severely from negative task interference.
When an agent attempts to learn conflicting objectives simultaneously, such as the aggressive high-speed tracking required for inspection versus the precise stabilization required for docking, the shared gradients often conflict, destabilizing the training process \cite{yu2020gradient}.

To mitigate this, optimization-based methods like Gradient Surgery (PCGrad) \cite{yu2020gradient} and recent task-specific action correction mechanisms \cite{feng2025efficient} attempt to project conflicting gradients or adjust actions to prevent destructive interference.

Alternatively, architectural approaches seek to compartmentalize task knowledge.
Soft Modularization \cite{yang2020multi} and Routing Networks \cite{devin2017learning} dynamically route inputs through different neural modules, while PaCo (Parameter-Compositional RL) learns a set of basis parameters to interpolate task policies \cite{sun2022paco}.
More recently, Mixture of Orthogonal Experts (MOORE) \cite{hendawy2023multi} and Projected Task-Specific Layers \cite{somerville2023projected} have been proposed to enforce orthogonality between task representations.

While these architectural methods reduce interference, they typically rely on learning discrete routing strategies or large quantities of expert parameters, which can be computationally heavy for flight hardware.
Furthermore, they often treat tasks as categorical indices.
In contrast, our work leverages Hypernetworks \cite{ha2016hypernetworks, schopf2022hypernetwork} to map a continuous, physics-informed semantic manifold directly to policy weights, enabling seamless behavioral interpolation without the complexity of mixture-of-experts routing.

\subsection{Hypernetworks and Context-Aware Control}
In RL, Hypernetworks have been widely adopted to address the challenges of generalization and memory stability. They are commonly used in Meta-RL \cite{beck2023hypernetworks} and Continual Learning \cite{schopf2022hypernetwork} to generate task-specific parameters that quickly adapt to new or sequential tasks.

In robotics, these modulation techniques are increasingly used to adapt policies to changing physics or capabilities.
For instance, context embeddings have been used to implicitly model varying system parameters such as friction or payload mass \cite{yang2020multi}.
Very recently, Hypernetworks have been effectively employed to handle heterogeneous robot capabilities in multi-agent teams \cite{fu2025capability} and combined with task-aware scene representations for robust manipulation \cite{sun2025hypertasr}.
However, a unified framework that leverages semantic task contexts to resolve conflicting GNC objectives in microgravity remains absent. Our work bridges this gap by proposing a Hypernetwork architecture that generalizes across diverse mission profiles and physical domains, suitable for the constraints of platforms like Int-Ball2 \cite{yamaguchi2024intball2}.

\section{Problem Formulation}

A single task $T_i$ in RL can be formulated as a Markov Decision Process (MDP) defined by the tuple $\mathcal{M}=(\mathcal{S,A},P,r,\gamma)$ where $\mathcal{S}$ is the state space, $\mathcal{A}$ is the action space, $P(s'|s,a)$ is the transition probability between states, $r(s,a)$ is the reward function, and $\gamma$ is the discount factor. In MTRL, the goal is to learn a policy $\pi$ that maximizes the expected return $J(\pi)$ for a set of $N$ tasks $\mathcal{T}=\{T_1,\dots,T_N\}$. The optimization objective is defined as
\begin{equation}
    J(\pi)=\frac{1}{N}\sum_{i=1}^N\mathbb{E}_\pi\left[\sum_{t=0}^\infty\gamma_i^tr_i(s_t,a_t)\right]
\end{equation}
where the expectation is taken with respect to the trajectory distribution induced by $\pi$ and the task-specific dynamics $P_i$.

\subsection{Robots' dynamics}
We model the Int-Ball2 as a 6-DOF free-flying robot based on the specifications provided by \citet{Mitani2023IntBall2}. The simulation environment captures the coupled translational and rotational dynamics of the eight-propeller configuration, enabling the development of control laws for high-precision maneuvers. For physical validation, we deploy these policies onto a satellite emulator consisting of a 5kg cylindrical floating platform with a 30cm radius. This platform utilizes an air-bearing system to emulate the planar microgravity dynamics of an orbital environment, allowing us to evaluate the policy's robustness against real-world friction, actuator latency, and mass distribution uncertainties.

\subsection{Tasks}\label{subsec:Tasks}
In this section, we present the tasks on which the robot is trained. To ensure reproducibility, the observation spaces are provided in the Appendix, \cref{tab:tasks_obs}, and all numerical constants and penalty terms are listed in \cref{table:reward_functions}.

\subsubsection{Track Velocities}\label{task:trackvelos} In the velocity tracking task, the robot must match a set of target linear and angular velocities across all six degrees of freedom. The reward function $r_t^{\text{track}}$ is defined as follows: $$r_t^{\text{track}} = r_t^{\text{lin}} + r_t^{\text{ang}}$$ where $r_t^{\text{lin}}$ and $r_t^{\text{ang}}$ reward the robot for minimizing the error between its current and target linear and angular velocity components, respectively.

\subsubsection{Docking} In the docking task, the robot must approach a target position and align its heading with the target direction. The reward function $r_t^{\text{docking}}$ is defined as follows: $$r_t^{\text{docking}} = r_t^{\text{pose}} + r_t^{v,\omega} + r_t^{\text{bnd}} + r^{\text{prog}}$$ where $r_t^{\text{pose}}$ rewards the robot for reaching the target position and heading simultaneously, $r_t^{v,\omega}$ encourages the robot to maintain linear and angular velocity within a desired range, $r_t^{\text{bnd}}$ penalizes the robot's proximity to a defined maximum distance boundary, keeping the robot focused within the work space, and $r^{\text{prog}}$ provides a positive incentive proportional to the reduction in distance from the goal over the last time step.

\subsubsection{Inspection} In the Inspection task, the robot must sequentially navigate through a pre-defined sequence of target poses. The reward function $r_t^{\text{insp}}$ is defined as: $$r_t^{\text{insp}} = r_t^{\text{pose}} + r_t^{\text{prog}} + r_t^{v,\omega} + r_t^{\text{bnd}} + r_t^{\text{seq}}$$ where $r_t^{\text{seq}}$ rewards the robot when it reaches a target and the other rewards are transferred from the previous task.

\subsubsection{Navigation with Obstacles} In the Navigation with Obstacles task, the robot seeks to reach a target position while avoiding collisions. The reward function $r_t^{\text{nav}}$ is defined as: $$r_t^{\text{nav}} = r_t^{\text{position}} + r_t^{v,\omega} + r_t^{\text{bnd}} + r_t^{\text{coll}}$$ where $r_t^{\text{position}}$ rewards the robot for reaching the target position, and $r_t^{\text{coll}}$ penalizes the robot when colliding with an obstacle.

\section{Methodology}

\begin{figure}
    \centering
    \includegraphics[width=0.49\textwidth]{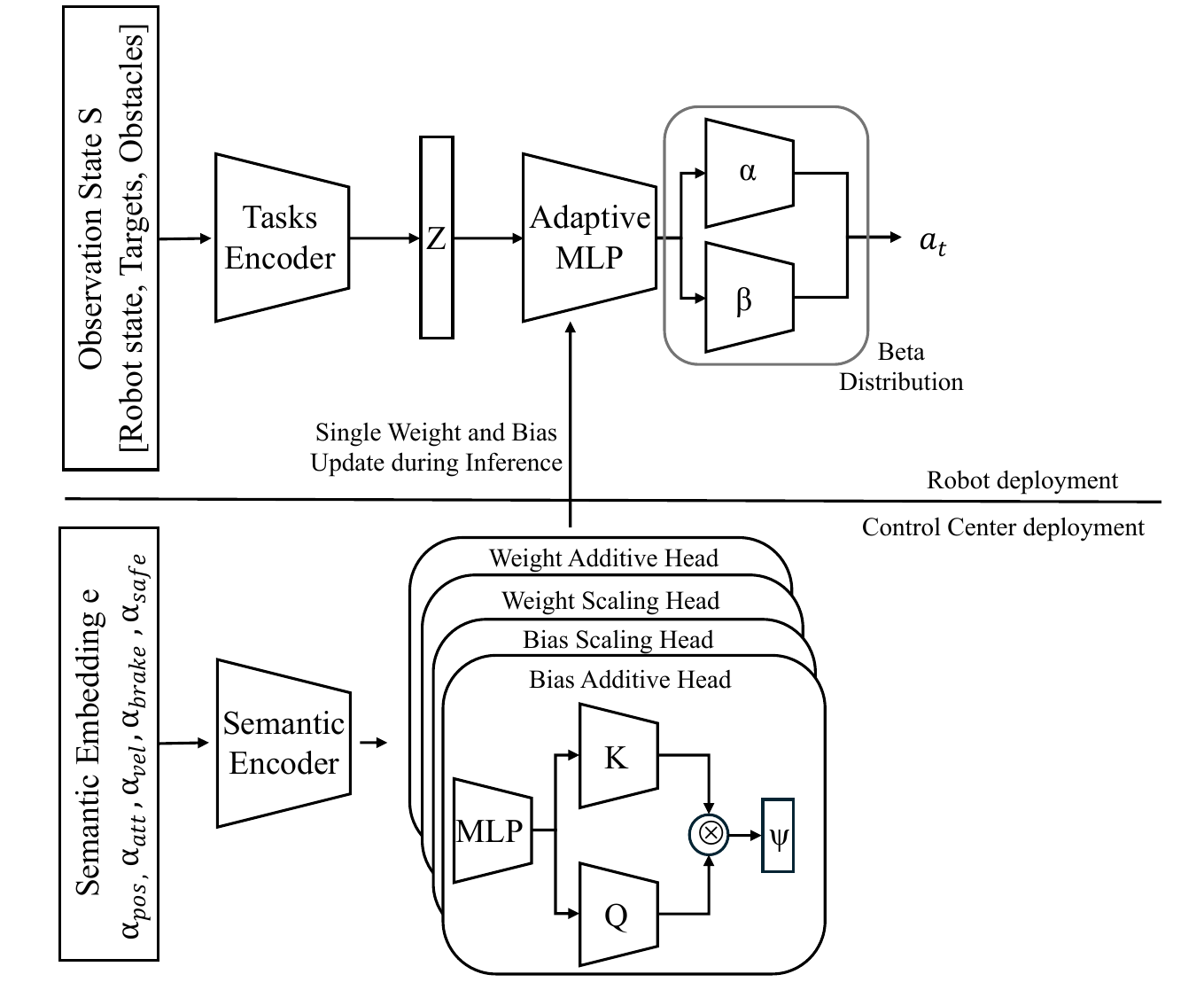}
    \caption{\textbf{Framework Overview.} We train a single model to perform multiple tasks. Depending on the objective, the weights of the model are modulated by the hypernetwork. The entire system is trained end-to-end using Proximal Policy Optimization (PPO) \cite{schulman2017proximalpolicyoptimizationalgorithms}.}
    \label{fig:framework_overview}
    \vspace{-1.5em}
\end{figure}

This section introduces \ours, an RL-based framework for learning a single policy that can adapt to multiple GNC tasks, as summarized in \cref{fig:framework_overview}. This approach enables a semantically rich understanding of each task and compositional synthesis of novel behavioral combinations. We first outline the key challenges addressed in this work in \cref{subsec:challenges}, then describe the core components of the framework in the following sections.

\subsection{Key Challenges \& Overview}\label{subsec:challenges}

\subsubsection{Conflicting Gradient Dynamics (\cref{subsec:conf_grad_dynam})} Spacecraft mission profiles require fundamentally opposing control laws. Docking requires soft dynamics, precise pose alignment, and aggressive damping (braking) near the target. In contrast, Inspection and Velocity Tracking require stiff tracking of reference vectors and the maintenance of momentum. A naive shared policy (e.g., standard PPO, multi-critic PPO) averages these gradients, resulting in a controller that is too aggressive for safe docking yet too slow for high-speed tracking.

\subsubsection{Behavioral Composition for Novel Mission Profiles (\cref{subseczero-shot-general})} It is computationally intractable to train a policy for every possible combination of mission requirements (e.g., Docking with Obstacles or Stabilization under failure). Standard MTRL approaches that use categorical Task IDs (e.g., one-hot vectors) cannot synthesize these novel behavioral combinations because the network treats tasks as discrete classes rather than points on a continuous behavioral manifold. This limitation directly diminishes operational versatility: any unanticipated scenario requires ground-based intervention to update the controller. We define \textit{zero-shot task composition} as the ability to combine learned behaviors at inference by constructing a novel embedding from known physical priorities, requiring zero retraining, reward specification, or gradient steps.

\subsubsection{Sim2Real Gap (\cref{subsec:sim2real})} Validating 3D orbital policies on ground-based testbeds creates a severe domain gap. We use the same setup as for Int-Ball2 to train a policy for a Floating Platform. Notably, while the Int-Ball2 utilizes a continuous action space, the Floating Platform operates on a binary action space. To accommodate this during sim-to-real transfer, we switch the policy's output from a Beta distribution to a Gaussian. A Beta distribution whose actions concentrate at the boundaries requires $\alpha_\theta(s) \to \infty$ or $\beta_\theta(s) \to \infty$, which produces gradient instability; a Gaussian avoids this and provides stable convergence when the output is discretized for binary thruster control \cite{tang2020discretizingcontinuousactionspace}. We deploy it in our laboratory and study the effect of domain randomization during training for the success of the tasks in the real world.

\subsection{Context-Aware Weight Modulation}\label{subsec:conf_grad_dynam}
\ours relies on a split architecture consisting of two primary components: (a) a Main Network, which processes robot observations, and (b) a Hypernetwork, which generates the parameters for the Main Network based on the task context.

\noindent \textbf{a) Main Network (Policy):} The Main Network functions as the actor-critic agent. The Actor and Critic are instantiated as separate networks but utilize an identical architecture consisting of a sequence of Adaptive Linear Layers. In this configuration, the weights and biases of each layer are dynamically modulated at inference time by a corresponding Hypernetwork (one per layer). The Actor features two output heads that parameterize a Beta distribution (via shape parameters $\alpha$ and $\beta$). The observation vectors are concatenated and zero-padded to a fixed maximum length before being fed into the network.

\noindent \textbf{b) Hypernetwork:} A Hypernetwork is attached to each adaptive layer to generate context-specific modulations. It takes a semantic task embedding $e_k \in [0, 1]^5$ from task $T_k$ as input. The embedding is first processed by a shared MLP (Semantic Encoder) and then fed into four distinct Modulation Heads: Weight Scaling ($\Psi_{\text{scale}}^W$), Weight Additive ($\Psi_{\text{add}}^W$), Bias Scaling ($\Psi_{\text{scale}}^b$), Bias Additive ($\Psi_{\text{add}}^b$). Each head utilizes a Rank-1 Vector Factorization mechanism. Instead of predicting full modulation matrices, the network predicts two vectors (Key $\mathbf{k}$ and Query $\mathbf{q}$). Their outer product generates the modulation matrix/vector, efficiently capturing task correlations.

\textbf{Stable initialization.} To avoid disrupting the pretrained base network at training onset, the Key/Query head weights are initialized with near-zero variance ($\sigma = 10^{-4}$) and the scale head output is shifted by $+1.0$, so that initially $\Psi_{\text{scale}} \to 1$ and $\Psi_{\text{add}} \to 0$, i.e., $\mathbf{W}_{\text{dyn}} \approx \mathbf{W}_{\text{base}}$. This is analogous to the residual $\gamma$ initialization used in ResNets \cite{he2018bagtricksimageclassification} and FiLM conditioning \cite{perez2017filmvisualreasoninggeneral}, and ensures the Hypernetwork begins as an identity modulation before task-specific weight adaptations emerge.

Let $z \in \mathbb{R}^{d_{in}}$ be the input to an adaptive layer and $\mathbf{W}_{\text{base}}$, $\mathbf{b}_{\text{base}}$ its static weights and biases. The layer computes the output $y \in \mathbb{R}^{d_{out}}$ using dynamic weights $\mathbf{W}_{\text{dyn}}$ and biases $\mathbf{b}_{\text{dyn}}$:

\begin{equation}
y = \mathbf{W}_{\text{dyn}}(e_k) z + \mathbf{b}_{\text{dyn}}(e_k)
\end{equation}

\noindent These dynamic parameters are derived from the base via a residual scale-and-shift modulation using batch multiplication.

\begin{equation}
\mathbf{W}_{\text{dyn}}(e_k) = \mathbf{W}_{\text{base}} \times (1 + \mathbf{\Psi}_{\text{scale}}^W(e_k)) + \mathbf{\Psi}_{\text{add}}^W(e_k)
\end{equation}
\begin{equation}
\mathbf{b}_{\text{dyn}}(e_k) = \mathbf{b}_{\text{base}} \times (1 + \mathbf{\Psi}_{\text{scale}}^b(e_k)) + \mathbf{\Psi}_{\text{add}}^b(e_k)
\end{equation}

\noindent To generate a modulation matrix $\Psi \in \mathbb{R}^{d_{out} \times d_{in}}$ efficiently, each hypernetwork head predicts two vectors: a Query vector $\mathbf{q}(e_k) \in \mathbb{R}^{d_{out}}$ and a Key vector $\mathbf{k}(e_k) \in \mathbb{R}^{d_{in}}$. The modulation matrix is computed as the outer product:

\begin{equation}
    \mathbf{\Psi}(e_k) = \frac{\mathbf{q}(e_k) \times \mathbf{k}(e_k)}{\sqrt{d_{\text{out}}}}
\end{equation}

\noindent \textbf{Rank-1 efficiency.} This factorization is motivated by parameter efficiency on flight hardware. A full-rank modulation head for a $[32 \times 32]$ layer requires predicting 1,024 values per head; with 4 heads per adaptive layer this yields 4,096 extra hypernetwork parameters per layer. Rank-1 prediction of key $\mathbf{k} \in \mathbb{R}^{32}$ and query $\mathbf{q} \in \mathbb{R}^{32}$ reduces this to 64 values---a $16\times$ reduction---while maintaining the full scale-and-shift expressivity of the modulation. This is analogous to LoRA \cite{hu2021loralowrankadaptationlarge}, where rank-1 updates have been shown sufficient for parameter-efficient adaptation. A more detailed description is available in the Appendix \cref{app:more_impl_details}.

\begin{figure*}[t]
    \centering
    \includegraphics[width=\textwidth]{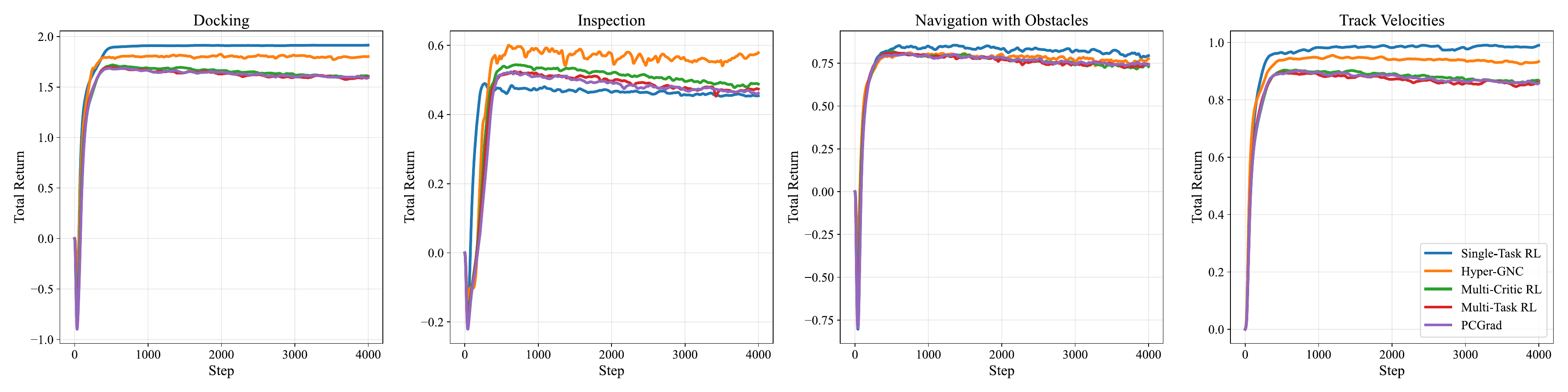}
    \caption{\textbf{Comparative Returns for Int-Ball2 GNC Tasks.} The proposed Hypernetwork architecture (Orange) demonstrates competitive sample efficiency, matching the convergence rate of Single-Task Experts (Blue) across all four mission profiles. }
    \label{fig:training_reward_curves}
    \vspace{-1em}
\end{figure*}

\subsection{Physics-Informed Hypernetwork Framework}\label{subseczero-shot-general}

To address these challenges, we propose a modular architecture that decouples task semantics from control execution. Our framework consists of a Semantic Task Encoder that modulates the weights of a shared Actor-Critic backbone via a Hypernetwork, see \cref{tab:sem_emb}.

Instead of discrete task IDs, we project all mission objectives into a continuous semantic manifold. We define the context vector $e_k \in [0, 1]^5$ as a set of normalized physical control priorities:

$$e_k = [\alpha_{pos}, \alpha_{att}, \alpha_{vel}, \alpha_{brake}, \alpha_{safe}]$$

\noindent where $\alpha_{pos}$ and $\alpha_{att}$ govern the position and attitude errors, $\alpha_{vel}$ dictates velocity maintenance, $\alpha_{brake}$ enables negative acceleration, and $\alpha_{safe}$ activates obstacle avoidance behaviors. This embedding is analogous to the GNC priority vector used in classical mission design \cite{wie2008space,fehse2003automated}: its five dimensions directly map to the distinct, independent control objectives of standard GNC mission decomposition rather than abstract learned features. During inference, we can synthesize unseen behaviors by interpolating these vectors (e.g., combining $\alpha_{brake}=1$ and $\alpha_{safe}=1$ for Safe Docking with Obstacles).

\noindent \textbf{Continuous sampling as regularization.} Each dimension is uniformly sampled from a range during training (see \cref{tab:sem_emb}), which prevents the policy from overfitting to discrete task IDs and acts as a form of parameter-space data augmentation. This explains why \ours outperforms \ours + TaskID: the categorical baseline treats tasks as isolated classes, producing brittle specialization, while continuous sampling learns weight modulations that generalize across the trained range.

\noindent \textbf{Deployment.} The semantic embedding $e_k$ can be set by a human operator, a higher-level mission planner, or hard-coded per mission phase---it requires no online learning or retraining.

\subsection{Sim-to-Real Transfer \& Domain Randomization}\label{subsec:sim2real}
We modify the Isaac Lab simulator \cite{mittal2025isaaclab} to train all four policies simultaneously in 4,096 parallel environments (1,024 per task), with PPO updates computed over all tasks in each iteration. More implementation details can be found in Appendix \cref{app:more_implementation_details}.

To facilitate robust sim-to-real transfer, we implement domain randomization \cite{8202133} by perturbing three key physical attributes: the center of mass (CoM) position, total mass variation, and the magnitude of external wrenches acting on the robot body.

\section{Simulation Experiments}

\begin{table*}[h!]
    \centering
    \setlength{\tabcolsep}{7pt}
    \renewcommand{\arraystretch}{1.2}
    \caption{\textsc{Int-Ball2 four GNC tasks comparison performance.}}
    \resizebox{\textwidth}{!}{
        \begin{tabular}{l *{2}{c}  *{2}{c} *{2}{c}  *{1}{c}}
        \toprule
        \multirow{3}{*}{\textbf{Methods}}
        & \multicolumn{2}{c}{\textit{Docking}} 
        & \multicolumn{2}{c}{\textit{Track Velocities}} 
        & \multicolumn{2}{c}{\textit{Inspection}} 
        & \multicolumn{1}{c}{\textit{Navigation with Obstacles}} \\
        & $e_p [m] \downarrow$ & $e_o [rad] \downarrow$
        & $e_v [m\,s^{-1}] \downarrow$ & $e_{\omega} [rad\,s^{-1}] \downarrow$
        & $e_o [rad] \downarrow$ & $SR [\%] \uparrow$
        & $e_p [m] \downarrow$ \\
        \midrule
        Single-Task RL  
            & 0.017 \ci{0.00773} & 0.041 \ci{0.02171}
            & 0.106 \ci{0.05330} & 0.178 \ci{0.06921}
            & 0.622 \ci{0.15115} & 85.46
            & 0.056 \ci{0.02538}\\
        \midrule
        Multi-Task RL 
            & 0.036  \ci{0.02008} & 0.126  \ci{0.06725}
            & 0.233  \ci{0.09422} & 0.386  \ci{0.03252}
            & 0.661  \ci{0.15632} & 62.50
            & 0.115  \ci{0.07620} \\
        Multi-Task RL TaskID
            & 0.018  \ci{0.00795} & 0.037  \ci{0.01281}
            & 0.183  \ci{0.10423} & 0.196  \ci{0.05987}
            & 0.578  \ci{0.16597} & 83.75
            & 0.043  \ci{0.02387} \\
        Multi-Task RL SemEmb
            & 0.025  \ci{0.00863} & 0.034  \ci{0.01461}
            & 0.180  \ci{0.06820} & 0.209  \ci{0.05861}
            & 0.626  \ci{0.15893} & 87.50
            & 0.025  \ci{0.00863} \\
        \midrule
        PCGrad  
            & 0.032  \ci{0.01351} & 0.136  \ci{0.06322}
            & 0.281  \ci{0.10991} & 0.403  \ci{0.03174}
            & \textbf{0.400  \ci{0.49144}} & 60.35
            & 0.113  \ci{0.09077} \\
        PCGrad HotID  
            & 0.022  \ci{0.01192} & 0.037  \ci{0.01720}
            & 0.177  \ci{0.07485} & 0.211  \ci{0.06390}
            & 0.605  \ci{0.17007} & 39.15
            & 0.035  \ci{0.02124} \\
        PCGrad SemID  
            & 0.019  \ci{0.00407} & 0.033  \ci{0.02254}
            & 0.179  \ci{0.07346} & 0.205  \ci{0.06196}
            & 0.633  \ci{0.18376} & 40.58
            & 0.041  \ci{0.02098} \\
        \midrule
        Multi-Critic  
            & 0.026  \ci{0.00477} & 0.046  \ci{0.02683}
            & 0.187  \ci{0.08791} & 0.246  \ci{0.06491}
            & 0.585  \ci{0.13618} & 85.00
            & 0.060  \ci{0.03419} \\
        Multi-Critic TaskID  
            & 0.020  \ci{0.00612} & 0.023  \ci{0.01933}
            & 0.150  \ci{0.07522} & 0.203  \ci{0.07368}
            & 0.621  \ci{0.17097} & 83.75
            & 0.096  \ci{0.76961} \\
        Multi-Critic SemID  
            & 0.023  \ci{0.01048} & 0.045  \ci{0.01719}
            & 0.198  \ci{0.09521} & 0.205  \ci{0.06566}
            & 0.595  \ci{0.17626} & 75.00
            & 0.049  \ci{0.02244} \\
        \midrule
        \ourrow \ours (ours)
            & \textbf{0.005  \ci{0.00258}} & \textbf{0.011  \ci{0.00711}}
            & \textbf{0.086  \ci{0.04103}} & \textbf{0.157  \ci{0.05246}}
            & 0.597  \ci{0.15920} &  \textbf{91.25}
            & \textbf{0.021  \ci{0.00891}} \\
        \bottomrule
        \end{tabular}
    }
    \label{tab:results_IntBall2}
    \vspace{-1em}
\end{table*}

Using the tasks described in the previous sections, we design our experiments to answer the following questions: (i) Can the \ours framework efficiently learn generalizable control policies across diverse GNC tasks? (ii) Can \ours synthesize novel behavioral combinations beyond its training task set to address evolving mission requirements? (iii) Does the proposed framework transfer successfully from simulation to a real-world robotic platform while preserving task performance?

\subsection{Experiment Setup}
In this section, we evaluate the performance of our methods. For Int-Ball2, we evaluate the model on the tasks presented in \cref{subsec:Tasks}. Then we test the model on novel behavioral combinations such as Docking with Obstacles, Point Navigation, and Stabilization. All the quantitative results are in \cref{tab:results_IntBall2} and  \cref{tab:OoD_results_IntBall2}.

\subsubsection{Track Velocities} We evaluate the performance of tracking linear and angular velocity vectors.

\subsubsection{Docking with and without obstacles}  Starting from random positions and orientations, we evaluate the final distance and orientation between the robot and the target goal.

\subsubsection{Inspection} We assess the orientation error between the robot and the target orientation and the success rate of reaching all the targets.

\subsubsection{Navigation with and without Obstacles} We measure the distance between the robot and the target navigation point.

\subsubsection{Stabilization} We initialize the system with random velocities and measure the final deviation from zero linear and angular velocity at the end of each episode.

\subsection{Baselines}
To evaluate the effectiveness of the key design choices in \ours, we compare it against state-of-the-art methods and ablated versions:
\begin{itemize}
  \item \textbf{Single-Task RL:} Fully retrained, task-specific specialists: a default PPO trained from scratch on a single task. For zero-shot task composition scenarios, this represents the performance upper bound achievable at the cost of full retraining.
  \item \textbf{Multi-Task RL (MTRL):} A baseline of concatenating all task information into a single observation vector with task IDs.
  \item \textbf{PCGrad:} Built on top of Multi-Task RL, it integrates the gradient surgery presented in \cite{yu2020gradientsurgerymultitasklearning}.
  \item \textbf{Multi-Critic (MTCR):} A baseline implementing PPO with a critic per task presented in \cite{mysore2022multicritic}.
\end{itemize}

\subsection{Main Results}

\ours demonstrates the best performance in learning the four GNC tasks, as shown in \cref{tab:results_IntBall2}. The effect of key design choices is summarized as follows:

\noindent \textbf{Training Efficiency (\cref{fig:training_reward_curves}).}  Illustrates the training progression across the four mission profiles. We observe that our Hypernetwork-based approach (Orange) demonstrates remarkable sample efficiency, achieving convergence rates comparable to the single-task Experts (Blue) and significantly outperforming standard multi-task baselines (MTRL, MTCR, PCGrad). Notably, the Hypernet reaches near-optimal policy performance within the same limited number of episodes as the specialized experts.

A distinct observation from our results is that while the Single-task Experts achieve marginally higher returns during training, the Hypernetwork consistently outperforms them in deployment evaluations, see \cref{tab:results_IntBall2}. We attribute this to the regularization effect inherent in our multi-task architecture. While the single-task Experts tend to overfit to specific training seeds or exploit reward function artifacts, maximizing return at the expense of robustness, the Hypernetwork is forced to learn a generalized control manifold shared across tasks. This results in a slightly lower peak training reward but yields a more robust policy that generalizes significantly better to the evaluation domain.

\begin{table}[b!]
    \centering
    \setlength{\tabcolsep}{7pt}
    \vspace{-1.5em}
    \renewcommand{\arraystretch}{1.2}
    \caption{\textsc{Int-Ball2 performance on composition tasks.}}
    \resizebox{0.49\textwidth}{!}{
        \begin{tabular}{l | *{2}{c} | *{2}{c} | *{1}{c}}
        \toprule
        \multirow{3}{*}{\textbf{Methods}}
        & \multicolumn{2}{c}{\textit{Docking with Obstacles}} 
        & \multicolumn{2}{c}{\textit{Stabilization}}
        & \multicolumn{1}{c}{\textit{Point Navigation}} \\
        & $e_p [m] \downarrow$ & $e_o [rad] \downarrow$
        & $e_v [m\,s^{-1}] \downarrow$ & $e_{\omega} [rad\,s^{-1}] \downarrow$
        & $e_p [m] \downarrow$\\
        \midrule
        Single-Task RL   
            & \textbf{0.057 \ci{0.07554}} & \textbf{0.141 \ci{0.09880}}
            & 0.001 \ci{0.00213} & 0.037 \ci{0.07641}
            & 0.019 \ci{0.00897}\\
        \midrule
        Multi-Task RL   
            & 0.079  \ci{0.23516} & 0.210  \ci{0.23673}
            & 0.425  \ci{0.17721} & 0.680  \ci{0.05634}
            & 0.037  \ci{0.02363} \\
        Multi-Task RL SemEmb
            & 0.081  \ci{0.18435} & 0.617  \ci{0.33866}
            & 0.028  \ci{0.01663} & 0.484  \ci{0.28859}
            & 0.036  \ci{0.02919} \\
        \midrule
        PCGrad     
            & 0.080  \ci{0.26632} & 0.210  \ci{0.24936}
            & 0.055  \ci{0.02401} & 0.952  \ci{0.39236}
            & 0.032  \ci{0.01433} \\
        PCGrad SemEmb        
            & 0.076  \ci{0.15134} & 0.799  \ci{0.54258}
            & 0.010  \ci{0.01670} & 0.110  \ci{0.16856}
            & 0.045  \ci{0.04843} \\
        \midrule
        Multi-Critic  
            & 0.092  \ci{0.29215} & 0.206  \ci{0.34149}
            & \textbf{0.000  \ci{0.00085}} & \textbf{0.001  \ci{0.02025}}
            & 0.026  \ci{0.00479} \\
        Multi-Critic SemEmb  
            & 0.076  \ci{0.21651} & 0.705  \ci{0.47769}
            & 0.003  \ci{0.00421} & 0.046  \ci{0.05770}
            & 0.041  \ci{0.01943} \\
        \midrule
        \ourrow \ours (ours)
            & 0.076  \ci{0.26234} & 0.182  \ci{0.23268}
            & 0.001  \ci{0.00088} & 0.017  \ci{0.01139}
            & 0.017  \ci{0.00984} \\
        \bottomrule
        \end{tabular}
    }
    \label{tab:OoD_results_IntBall2}
    \vspace{-1em}
\end{table}

\noindent \textbf{Semantic Embeddings and Behavioral Composition.} While discrete Task IDs restrict a policy to a fixed set of known maneuvers, the use of a continuous semantic manifold enables the model to synthesize novel behavioral combinations. As seen in \cref{tab:OoD_results_IntBall2}, our method achieves competitive performance across all novel mission configurations. More importantly, the semantic manifold provides a framework for behavioral interpolation; by representing tasks as physical vectors, the Hypernetwork can modulate the policy for novel combinations of objectives (like docking in cluttered environments) without the need for task-specific retraining—directly supporting extended operational mission lifetimes.

\noindent \textbf{Beta distribution over Gaussian.} We choose a Beta distribution over a Gaussian because it is better suited for continuous control with a bounded action space \cite{petrazzini2021proximalpolicyoptimizationcontinuous}. Quantitative results in \cref{tab:apx_abl_and_results_IntBall2} strengthen this conclusion, with our approach achieving a 50.97\% improvement across all tasks over the Gaussian method.

\noindent \textbf{Unneeded task identifier on the observations.} While Multi-Critic and PCGrad methods benefit from task identifiers, such as one-hot encodings or semantic embeddings, \ours method remains agnostic to them, as the necessary task information is already captured by the Hypernetwork. As shown in \cref{tab:apx_abl_and_results_IntBall2}, the inclusion of redundant task identifiers actually results in a slight performance degradation.

\begin{figure}[h]
    \centering
    \includegraphics[width=0.5\textwidth]{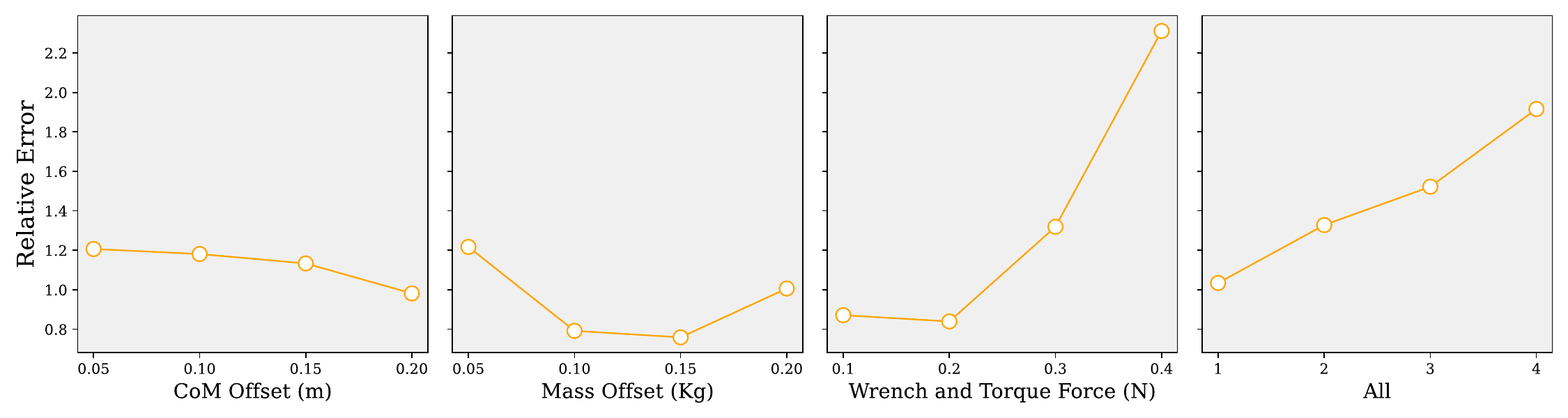}
    \caption{\textbf{Floating Platform robustness analysis in simulation.} A relative error of $1.0$ represents the performance of the non-randomized baseline; values above $1.0$ indicate performance degradation under perturbation.}
    \label{fig:robustness_analysis}
    \vspace{-1em}
\end{figure}

\begin{table*}[h!]
    \centering
    \setlength{\tabcolsep}{7pt}
    \renewcommand{\arraystretch}{1.2}
    \caption{\textsc{\textbf{Sim-to-Real results.}} Performance of \ours with and without domain randomization in the real world.}
    \resizebox{\textwidth}{!}{
        \begin{tabular}{l *{2}{c}  *{2}{c} *{2}{c}  *{1}{c} *{2}{c}  *{1}{c}}
        \toprule
        \multirow{3}{*}{\textbf{Methods}}
        & \multicolumn{2}{c}{\textit{Docking}} 
        & \multicolumn{2}{c}{\textit{Track Velocities}} 
        & \multicolumn{2}{c}{\textit{Inspection}} 
        & \multicolumn{1}{c}{\textit{Navigation with Obstacles}} 
        & \multicolumn{2}{c}{\textit{Stabilization}} 
        & \multicolumn{1}{c}{\textit{Point Navigation}} \\
        & $e_p [m] \downarrow$ & $e_o [rad] \downarrow$
        & $e_v [m\,s^{-1}] \downarrow$ & $e_{\omega} [rad\,s^{-1}] \downarrow$
        & $e_o [rad] \downarrow$ & $SR [\%] \uparrow$
        & $e_p [m] \downarrow$
        & $e_v [m\,s^{-1}] \downarrow$ & $e_{\omega} [rad\,s^{-1}] \downarrow$
        & $e_p [m] \downarrow$ \\
        \midrule
        \ours w/o Rand
            & 0.092 \ci{0.03890067482756609} & 0.198 \ci{0.17232992720422818}
            & 0.183 \ci{0.02860100736481281} & \textbf{0.152} \ci{0.03336737517314979}
            & 0.125 \ci{0.0064} & \textbf{100}
            & 2.136 \ci{0.6363179438084313}
            & 0.135 \ci{0.12478323067574923} & 0.155 \ci{0.14678131488749618}
            & 0.093 \ci{0.00510301791645885} \\
        \ours  
            & \textbf{0.028} \ci{0.019020509360028515} & \textbf{0.128} \ci{0.010986971392837287}
            & \textbf{0.169} \ci{0.015200476467327355} & 0.157 \ci{0.10748932221008088}
            & \textbf{0.082} \ci{0.0214} & \textbf{100}
            &  \textbf{0.087} \ci{0.005779587310422781} 
            & \textbf{0.077} \ci{0.0431498266202771} & \textbf{0.110} \ci{0.06557231077992505}
            & \textbf{0.094} \ci{0.0041839002416764245}\\
        \bottomrule
        \end{tabular}
    }
    \label{tab:sim2real_results}
    \vspace{-1em}
\end{table*}

\begin{figure}[h]
    \centering
    \includegraphics[width=0.5\textwidth]{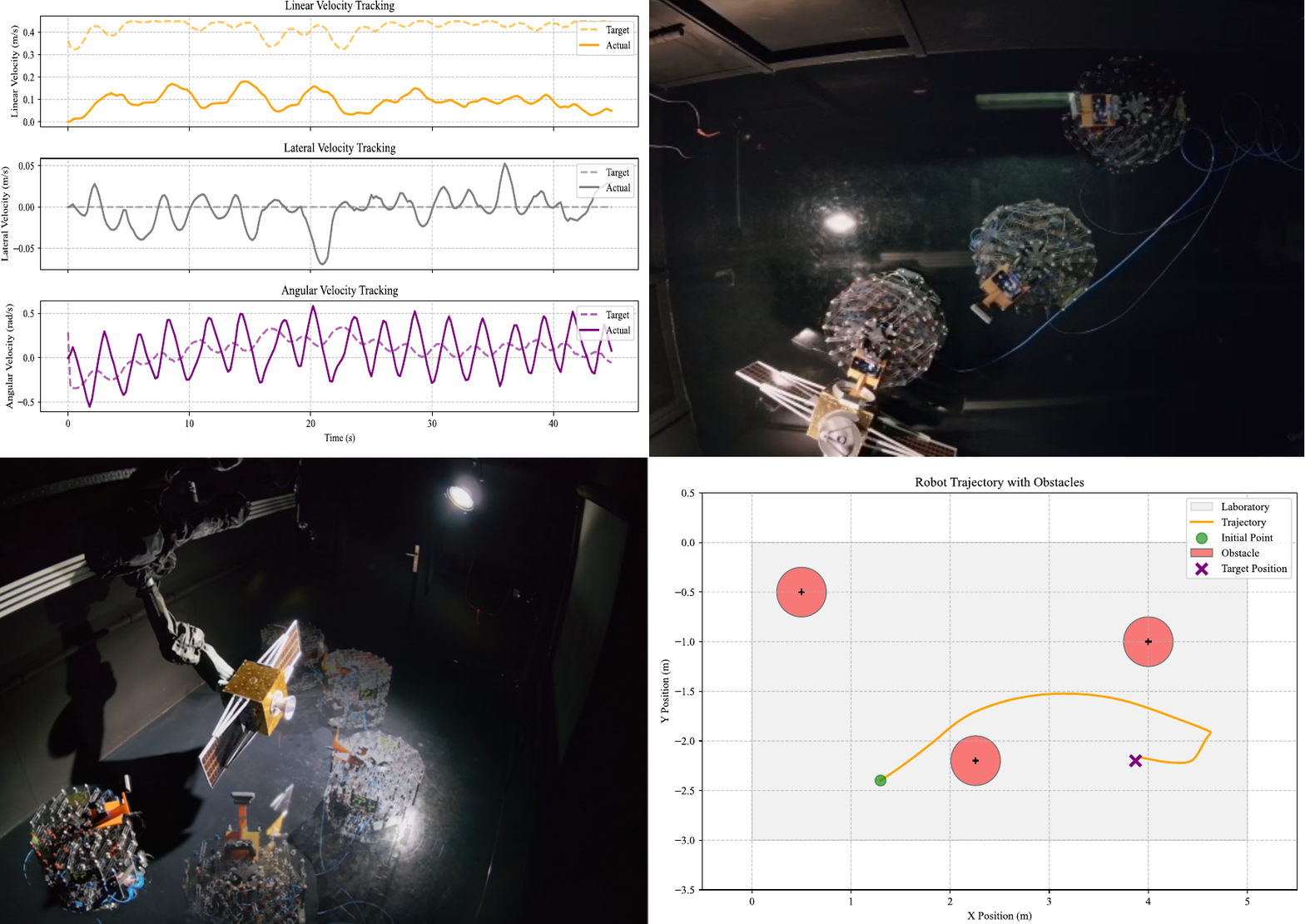}
    \caption{\textbf{\ours in the real world.} From left to right, top to bottom, target and actual velocities of the Track Velocities task, docking into a satellite, inspection of a satellite, and navigation through obstacles.}
    \label{fig:single_runs_vis}
    \vspace{-1em}
\end{figure}

\subsection{More Analyses}

\noindent \textbf{Combining \ours with MTCR and PCGrad (\cref{tab:apx_abl_and_results_IntBall2}).} We evaluated the compatibility of \ours with auxiliary multi-task optimization techniques. Surprisingly, the addition of PCGrad or Multi-Critic heads often resulted in performance degradation compared to the standalone \ours model. When both PCGrad and MTCR were applied concurrently, a slight performance lift was observed; however, this hybrid configuration failed to provide meaningful gains in behavioral composition and significantly increased computational overhead, see \cref{tab:training_times}. These findings imply that the inductive bias provided by \ours effectively manages task-specific nuances, making the standalone architecture the most robust and computationally efficient choice for real-time GNC deployment.

\noindent \textbf{Curriculum Learning.} We investigate the trade-off regarding obstacle density during the training phase. Implementing a curriculum strategy consistently enhances performance across all tasks. Empirically, we observe that a sigmoid-based difficulty scheduler facilitates policy convergence. This is likely because the sigmoid profile provides an initial period of stability to master basic task dynamics, followed by a graduated increase in environmental complexity during the mid-stages of training, finally stabilizing at the maximum obstacle density.

\noindent \textbf{Robustness analysis (\cref{fig:robustness_analysis})} We calculate the ratio between the performance of the policy trained without randomizations and evaluate it against four categories of environmental disturbances: center of mass (CoM) offset, mass offset, external wrench and torque forces, and a combined category with all disturbances. We observe that the controller is remarkably resilient to internal inertial changes, with the relative error remaining near a factor of $1.2$ and below. However, the system exhibits higher sensitivity to external disturbances, where wrench and torque forces exceeding $0.3$N result in a sharper non-linear increase in relative error. When all disturbances are applied simultaneously, the relative error scales linearly with the severity of the combined increments, demonstrating a graceful degradation of performance.

\section{Real Robot Experiments}

\subsection{Main Results}
We evaluated our method across the primary mission profiles: Docking, Velocity Tracking, Inspection, and Navigation with Obstacles, alongside novel behavioral combinations such as Stabilization and Point Navigation. We conducted three individual runs for
each task, varying the starting conditions in each run. Quantitative results, summarized in \cref{tab:sim2real_results}, demonstrate that the \ours framework successfully transfers to physical hardware.

\subsection{Sim-to-real Analysis}
Based on the robustness analysis described in \cref{fig:robustness_analysis}, we deploy a policy trained with $\mathcal{U}(0.0, 0.15)$ mass and CoM offsets and $\mathcal{U}(0.0, 0.2)$ wrench and torque forces. Our results indicate that incorporating these randomization terms significantly narrows the gap, particularly for the Navigation with Obstacles and Docking tasks. Empirically, we observe a high degree of smoothness in the generated trajectories. This is primarily attributed to the framework's ability to learn an efficient thruster activation scheme. \ours\ produces sparse, purposeful firings that minimize fuel consumption while maintaining precise attitude control. This efficiency suggests that the latent task embeddings effectively regularize the policy, favoring energy-optimal maneuvers over erratic corrections; see \cref{fig:actions_fp}.

\subsection{Sensitivity analysis}
\begin{figure}[h]
    \centering
    \includegraphics[width=0.5\textwidth]{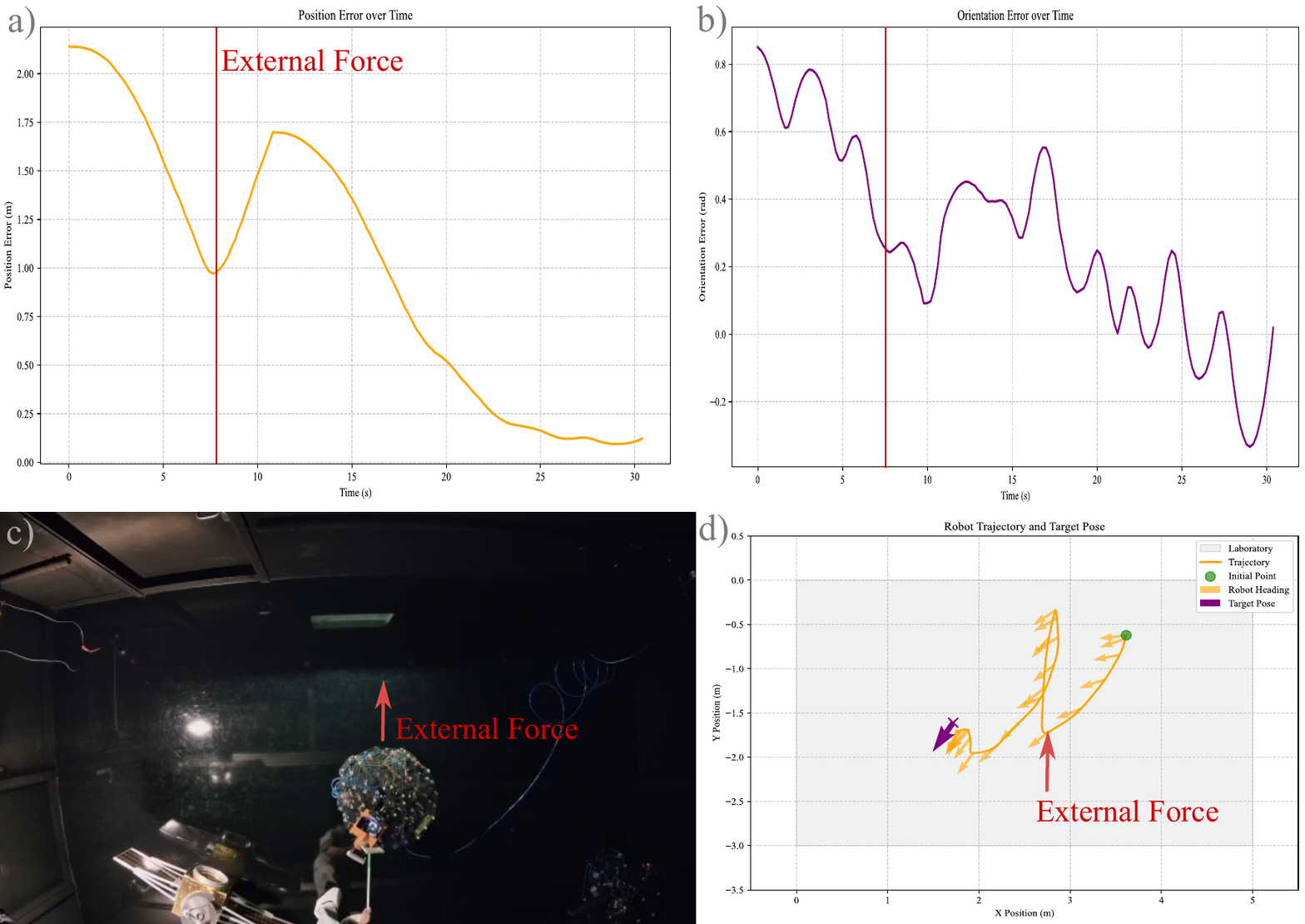}
    \caption{\textbf{Sensitivity analysis.} Single run of an external force applied during the docking task. a) Position error. b) Orientation error. c) Visual of the moment of contact. d) Full trajectory with orientation.}
    \label{fig:push_docking_sensitivity_analysis}
    \vspace{-2em}
\end{figure}

We perform a sensitivity analysis on the Docking task to check the performance when changing the mass, the center of mass with added weight (\cref{tab:sensitivity_analysis}), and external forces (\cref{fig:push_docking_sensitivity_analysis}).

\begingroup
\setlength{\tabcolsep}{4pt}
\begin{table}[h]
    \centering
    
    \resizebox{0.96\linewidth}{!}{%
 \begin{tabular}{lc c ccc c ccc} 
 \toprule
  \multirow{2}{*}{Metric} & & \multicolumn{3}{c}{Payload Mass} & & \multicolumn{3}{c}{CoM w/ 2kg} \\ 
   \cmidrule{3-5} \cmidrule{7-9}
    & 
    & 1kg &  2kg & 3kg  & 
    & 10cm &  20cm & 30cm  
    \\ 
 \midrule 
  $e_{\text{p}}[m]\downarrow$ & 
  & 0.143 \ci{0.177} & 0.059 \ci{0.043} & 0.107 \ci{0.093} &
  & 0.116 \ci{0.016} & 0.262 \ci{0.258} & 0.218 \ci{0.069} &
 \\[0.4ex]
 $e_{\text{o}}[rad]\downarrow$ & 
 & 0.204 \ci{0.119} & 0.177 \ci{0.154} & 0.176 \ci{0.145} &
 & 0.080 \ci{0.016} & 0.179 \ci{0.081} & 0.079 \ci{0.028} &
 
 \\
\bottomrule
\end{tabular}}
\label{tab:sensitivity_analysis}
\caption{\textsc{\textbf{Sensitivity Analysis.}} Robustness to payload, increasing the mass by 1kg, 2kg, and 3kg, and change of CoM with an offset of 10cm, 20cm, and 30cm using a 2kg added weight.}
\vspace{-1em}
\end{table}
\endgroup

\section{Conclusion}

This paper introduced \ours, a novel MTRL framework designed to provide free-flying robots with a unified, low-level control policy. By mapping a continuous semantic task manifold through a hypernetwork architecture, \ours successfully overcomes the interference typically encountered when training a single agent for heterogeneous mission objectives. We demonstrated that our framework can simultaneously master four distinct mission profiles, \textit{Docking}, \textit{Velocity Tracking}, \textit{Inspection}, and \textit{Obstacle Avoidance}, achieving sample efficiency comparable to single-task experts while maintaining behavioral composition for novel mission configurations.

Extensive evaluations with Int-Ball2 (simulation) and Floating Platform (real-world) validate the robustness and efficiency of our approach in real-world settings. Our results show that the learned manifold allows for control under physical perturbations and seamless switching between mission-critical tasks. Looking forward, this work provides a scalable foundation for modular free-flyers. This will enable autonomous systems to meet shifting mission requirements using a single, resource-efficient controller, ensuring long-term operational viability without ground intervention.

\section{Limitations and Future Directions}

While our method demonstrates strong performance across diverse GNC mission profiles and in real-world settings, we acknowledge several key limitations that should be addressed in the near future.

\noindent \textbf{Perception of the environment.} Although the main focus of the work is a low-level controller and most of the observations are sufficient with proprioception alone, tasks with more complex profiles, such as obstacle avoidance, assume perfect information about the environment. Integrating perceptual capabilities will help make the framework more complete.

\noindent \textbf{More diverse tasks.} The current task suite focuses on 6-DOF rigid body control. However, future orbital missions, such as active debris removal or on-orbit servicing, will require the coordination of robotic manipulators. We aim to explore how our Hypernetwork architecture scales when generating weights for policies governing coupled multi-body dynamics, where the task manifold must account for both base mobility and end-effector precision.

\noindent \textbf{Integration with Int-Ball2.} While our results include high-fidelity simulations of the Int-Ball2 platform, physical deployment has currently been limited to our laboratory satellite emulator.

\bibliographystyle{plainnat}
\bibliography{references}

\appendix\label{app}
\subsection{More Experimental Details}\label{app:more_implementation_details}
\noindent \textbf{Hardware Setup.} We train our policies on a single Nvidia RTX 4090. The Floating Platform has a mass of 5kg, a height of 0.5m, radius of 30cm, and 8 binary thruster actuation. We modeled the Int-Ball2 following \cite{Mitani2023IntBall2}.

\noindent \textbf{Evaluation.} Each policy is trained on 5 seeds 4000 iterations and each seed is evaluated on 128 episodes totaling 640 episodes per policy. The initial conditions are randomized. For Docking the initial robot location is sampled from $[-5,-0.5], [0.5, 5]\text{m}$ from the target position, yaw angle between $[-\pi, \pi]$ and a pitch, and roll angle between $[\frac{-\pi}{2}, \frac{\pi}{2}]$. For Track Velocities the initial linear velocity is sampled between $[0,2]$ m/s and the initial angular velocity is sampled between $[0,1.5]$ rad/s. For Inspection the initial robot location is sampled from $[-5,-0.5], [0.5, 5]\text{m}$ from the first target position, and for Navigation with Obstacles the initial robot location is sampled from $[-5,-3.5], [3.5, 5]\text{m}$ from the target position.

\subsection{More Implementation Details}\label{app:more_impl_details}
\noindent \textbf{Curriculum Setup.} The curriculum adjustment for the number of obstacles ($n$) follows the following function: $$n=\frac{n_{\text{max}}-n_{\text{min}}}{1+e^{-7(x-0.5)}}+n_{\text{min}}$$ where $x$ is the normalized step.

\noindent \textbf{PPO Implementation.} We adapt the PPO implementation of \citet{schwarke2025rslrl}. All actors and critics networks consist of 2-layer MLP of [32, 32] and [512, 512] neurons, respectively. All MLP in the Hypernetwork have 32 neurons. Each iteration includes 16 steps per environment, 5 learning epochs and 4 mini-batches per epoch. The discount factor $\gamma$ is set to 0.99, the clip ratio is set to 0.2, and the entropy coefficient is 0.005. We open-source our Hypernetwork implementation in our code repository.

\begin{table}[H]
    \centering
    \caption{\textsc{\textbf{Semantic Embedding Selection.}} To construct a continuous manifold, semantic embedding values are uniformly sampled for each task.}
    \renewcommand{\arraystretch}{1.2}
    \resizebox{1\linewidth}{!}{
    \begin{tabular}{l l }
    \toprule
    Task & Semantic Embedding \\
    \midrule

    Navigation with Obstacles &
        \begin{tabular}[t]{@{}l@{}}
            $\alpha_{pos}, \alpha_{brake}, \alpha_{safe} \in \mathcal{U}(0.8,1.0)$ \\
        \end{tabular} \\

        \addlinespace[3pt]

    Docking &
        \begin{tabular}[t]{@{}l@{}}
            $\alpha_{pos}, \alpha_{att}, \alpha_{brake} \in \mathcal{U}(0.8,1.0)$ \\
        \end{tabular} \\

        \addlinespace[3pt]

    Inspection &
        \begin{tabular}[t]{@{}l@{}}
            $\alpha_{pos} \in \mathcal{U}(0.7,1.0)$ \\
            $\alpha_{att}, \alpha_{vel} \in \mathcal{U}(0.8,1.0)$ \\
        \end{tabular} \\

        \addlinespace[3pt]

    Track Velocities &
        \begin{tabular}[t]{@{}l@{}}
            $\alpha_{vel} \in \mathcal{U}(0.8,1.0)$ \\
        \end{tabular} \\

        \addlinespace[3pt]

    Docking with obstacles&
        \begin{tabular}[t]{@{}l@{}}
            $\alpha_{pos}, \alpha_{att}, \alpha_{brake}, \alpha_{safe} \in \mathcal{U}(0.8,1.0)$ \\
        \end{tabular} \\

    Stabilization &
        \begin{tabular}[t]{@{}l@{}}
            $\alpha_{pos}, \alpha_{brake}, \alpha_{vel} \in \mathcal{U}(0.8,1.0)$ \\
        \end{tabular} \\

        \addlinespace[3pt]

    Navigation to a point &
        \begin{tabular}[t]{@{}l@{}}
            $\alpha_{pos}, \alpha_{brake} \in \mathcal{U}(0.8,1.0)$ \\
        \end{tabular} \\

    \bottomrule
    \end{tabular}}
    \label{tab:sem_emb}
\end{table}

\begin{table}[H]
    \centering
    \caption{\textsc{\textbf{Summary of Navigation Tasks and General Observation Space.}} In all tasks, the previous action is concatenated to the observation. Here $v_x$, $v_y$, $\omega_z$ are linear and angular velocities, $d$ is the Euclidean distance between the target and the robot, $\theta$ is the angle between the robot heading and the target location, $\psi$ is the angle between the target heading and the robot heading, $d_o$ is the Euclidean distance to an obstacle, and $o$ is the relative direction (bearing) to that obstacle.}
    \renewcommand{\arraystretch}{1.2} 
    \resizebox{1\linewidth}{!}{
    \begin{tabular}{l c l l}
    \toprule 
    Task & Dimension  & Variables & Components \\
    \midrule 
    
    \ourrow \multicolumn{4}{l}{\textbf{(a) Int-Ball2}} \\ 
    \noalign{\vskip 0.4mm}\cdashline{1-4}\noalign{\vskip 0.8mm}
    Navigation with Obstacles & $16 + 4n$ &
        \begin{tabular}[t]{@{}l@{}} 
            $[v_x, v_y, v_z, \omega_x, \omega_y, \omega_z]$ \\ 
            $[\Delta_{p_x}, \Delta_{p_y}, \Delta_{p_z}, \Delta_q]$ \\
            $[\mathds{1}]$ \\
            $n \times [d, o]$
        \end{tabular} & 
        \begin{tabular}[t]{@{}l@{}} 
            Base velocities \\ 
            Target information \\
            Collision signal \\
            Obstacles information \\
        \end{tabular} \\

    \addlinespace[3pt]

    Inspection & $15 + 9n$ & 
        \begin{tabular}[t]{@{}l@{}} 
            $[v_x, v_y, v_z, \omega_x, \omega_y, \omega_z]$ \\ 
            $[\Delta_{p_x}, \Delta_{p_y}, \Delta_{p_z}, \Delta_q]$ \\
            $n \times [\Delta_{p_{i_x}}, \Delta_{p_{i_y}}, \Delta_{p_{i_z}}, \Delta_{i_q}]$ \\
        \end{tabular} & 
        \begin{tabular}[t]{@{}l@{}} 
            Base velocities \\ 
            Target information \\
            Subsequent targets information \\
        \end{tabular} \\

    \addlinespace[3pt]
    
    Track Velocities & $12$ &
        \begin{tabular}[t]{@{}l@{}} 
            $[v_x, v_y, v_z, \omega_x, \omega_y, \omega_z]$ \\ 
            $[e_{\dot{x}}, e_{\dot{y}}, e_{\dot{z}}, e_{\dot{\theta}}, e_{\dot{\phi}}, e_{\dot{\psi}}, ]$ \\
        \end{tabular} & 
        \begin{tabular}[t]{@{}l@{}} 
            Base velocities \\ 
            Target information \\
        \end{tabular} \\

    \addlinespace[3pt]
        
    Docking & $15$ & 
        \begin{tabular}[t]{@{}l@{}} 
            $[v_x, v_y, v_z, \omega_x, \omega_y, \omega_z]$ \\ 
            $[\Delta_{p_x}, \Delta_{p_y}, \Delta_{p_z}, \Delta_q]$ \\
        \end{tabular} & 
        \begin{tabular}[t]{@{}l@{}} 
            Base velocities \\ 
            Target information \\
        \end{tabular} \\

    \midrule
    \ourrow \multicolumn{4}{l}{\textbf{(b) Floating Platform}} \\ 
    \noalign{\vskip 0.4mm}\cdashline{1-4}\noalign{\vskip 0.8mm}
    Navigation with Obstacles & $6+3n$ & 
        \begin{tabular}[t]{@{}l@{}} 
            $[v_x, v_y, \omega_z]$ \\ 
            $[d, \cos(\theta), \sin(\theta)]$ \\
            $n \times [d_{\text{o}}, cos(\eta), sin(\eta)]$
        \end{tabular} & 
        \begin{tabular}[t]{@{}l@{}} 
            Base velocities \\ 
            Target information \\
            Obstacle information
        \end{tabular} \\

    \addlinespace[3pt]
        
    Inspection & $8+5n$ &
        \begin{tabular}[t]{@{}l@{}} 
            $[v_x, v_y, \omega_z]$ \\ 
            $[d, \cos(\theta), \sin(\theta), \cos(\phi), \sin(\phi)]$ \\
            $n \times [d_{\text{i}}, cos(\theta), sin(\theta), \cos(\phi), \sin(\phi)]$
        \end{tabular} & 
        \begin{tabular}[t]{@{}l@{}} 
            Base velocities \\ 
            Target information \\
            Subsequent targets information
        \end{tabular} \\

    \addlinespace[3pt]
        
    Track Velocities & $6$ &
        \begin{tabular}[t]{@{}l@{}} 
            $[v_x, v_y, \omega_z]$ \\ 
            $[e_{\dot{x}}, e_{\dot{y}}, e_{\dot{\theta}},]$ 
        \end{tabular} & 
        \begin{tabular}[t]{@{}l@{}} 
            Base velocities \\ 
            Target information 
        \end{tabular} \\

    \addlinespace[3pt]
        
    Docking & $8$ & 
        \begin{tabular}[t]{@{}l@{}} 
            $[v_x, v_y, \omega_z]$ \\ 
            $[d, \cos(\theta), \sin(\theta), \cos(\phi), \sin(\phi)]$ \\
        \end{tabular} & 
        \begin{tabular}[t]{@{}l@{}} 
            Base velocities \\ 
            Target information \\
        \end{tabular} \\

    \bottomrule
    \end{tabular}}
    \label{tab:tasks_obs} 
\end{table}

\begin{table}[H]
    \centering
    \caption{\textsc{\textbf{Compute times and parameters.}} Training and evaluation average times for all methods, 5 seeds, on a single commercial GPU. The parameters of the Single-Task RL are ordered by Docking, Navigation with Obstacles, Inspection, and Track Velocities}
    \renewcommand{\arraystretch}{1.2}
    \resizebox{1\linewidth}{!}{
        \begin{tabular}{l r r}
        \toprule
        Method & Training Time [min] & Num. Parameters \\
        \midrule

        Single-Task RL & 783 & 74289 + 78897 + 79473 + 73425 \\
        Multi-Task RL & 291 & 79473 \\
        PCGrad & 528 & 79473 \\
        Multi-Critic & 428 & 309876 \\
        \ours & 423 & 128737 \\
        \ours MTCR & 437 & 320926 \\
        \ours PCGrad & 641 & 128737 \\
        \ours MTCR PCGrad & 707 & 320926\\

        \bottomrule
        \end{tabular}
    }
    \label{tab:training_times}

\end{table}

\begin{figure}[H]
    \centering
    \caption{\textbf{Navigation with Obstacles actions.} Individual thruster actions for a single navigation with obstacles run. \ours is able to learn policies that are fuel efficient.}
    \includegraphics[width=0.5\textwidth]{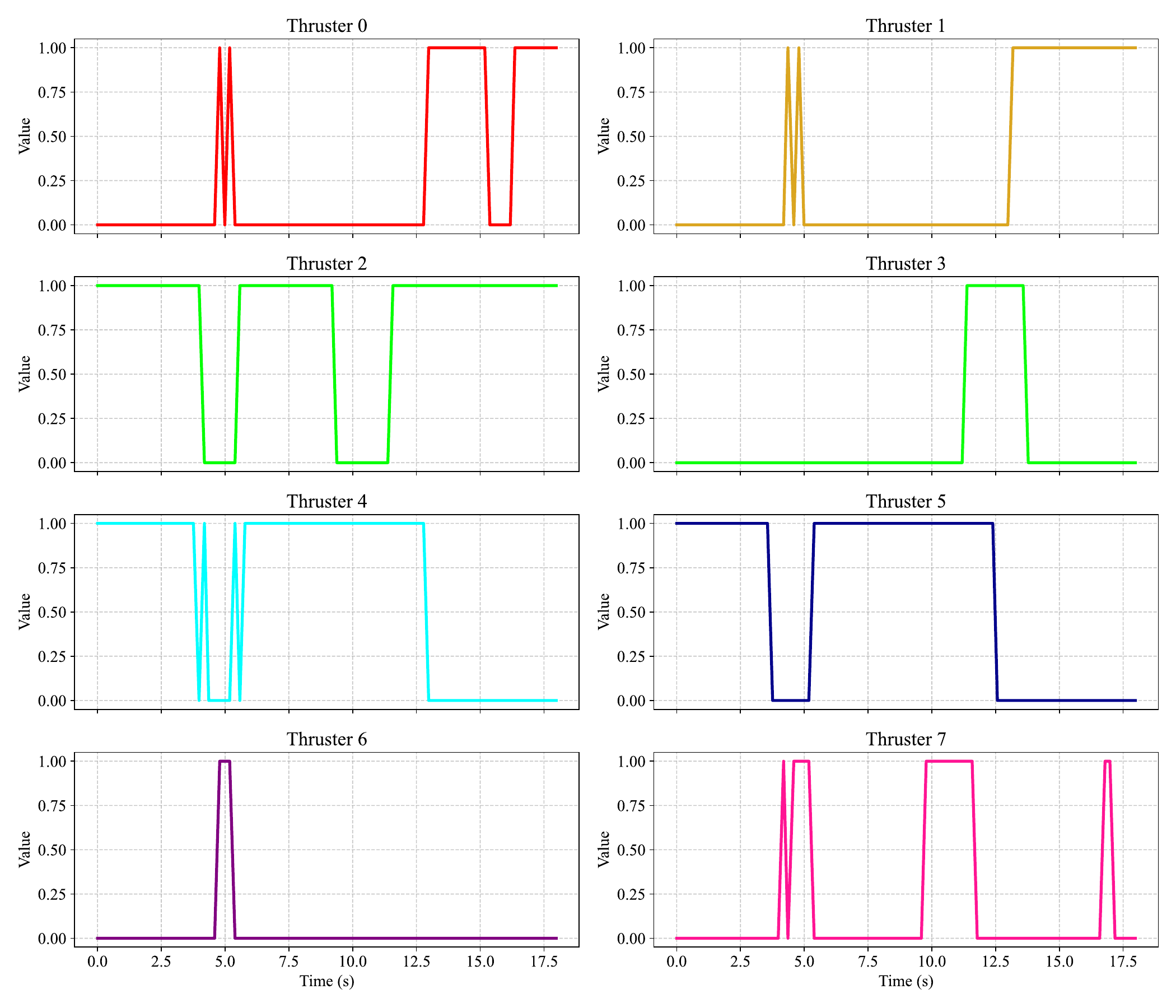}
    \label{fig:actions_fp}
\end{figure}

\begin{table}[H]
    \centering
    \setlength{\tabcolsep}{7pt}
    \renewcommand{\arraystretch}{1.2}
    \caption{\textsc{Performance on novel composition tasks with the same ablations of \cref{tab:apx_abl_and_results_IntBall2}.}}
    \resizebox{0.5\textwidth}{!}{
        \begin{tabular}{l | *{2}{c} | *{2}{c} | *{1}{c}}
        \toprule
        \multirow{3}{*}{\textbf{Methods}}
        & \multicolumn{2}{c}{\textit{Docking with Obstacles}}
        & \multicolumn{2}{c}{\textit{Stabilization}}
        & \multicolumn{1}{c}{\textit{Point Navigation}} \\
        & $e_p [m] \downarrow$ & $e_o [rad] \downarrow$
        & $e_v [m\,s^{-1}] \downarrow$ & $e_{\omega} [rad\,s^{-1}] \downarrow$
        & $e_p [m] \downarrow$\\
        \midrule
        Single-Task RL
            & 0.057 \ci{0.07554} & 0.141 \ci{0.09880}
            & 0.001 \ci{0.00213} & 0.037 \ci{0.07641}
            & 0.01900 \ci{0.00897}\\
        \midrule
        Multi-Task RL
            & 0.07974  \ci{0.23516} & 0.21004  \ci{0.23673}
            & 0.42543  \ci{0.17721} & 0.68095  \ci{0.05634}
            & 0.03744  \ci{0.02363} \\
        Multi-Task RL SemEmb
            & 0.08139  \ci{0.18435} & 0.61795  \ci{0.33866}
            & 0.02842  \ci{0.01663} & 0.48482  \ci{0.28859}
            & 0.03648  \ci{0.02919} \\
        \midrule
        PCGrad
            & 0.08076  \ci{0.26632} & 0.21088  \ci{0.24936}
            & 0.05543  \ci{0.02401} & 0.95217  \ci{0.39236}
            & 0.03266  \ci{0.01433} \\
        PCGrad SemID
            & 0.07647  \ci{0.15134} & 0.79912  \ci{0.54258}
            & 0.01075  \ci{0.01670} & 0.11098  \ci{0.16856}
            & 0.04567  \ci{0.04843} \\
        \midrule
        Multi-Critic
            & 0.09228  \ci{0.29215} & 0.20645  \ci{0.34149}
            & 0.00010  \ci{0.00085} & 0.00091  \ci{0.02025}
            & 0.02650  \ci{0.00479} \\
        Multi-Critic SemID
            & 0.07616  \ci{0.21651} & 0.70527  \ci{0.47769}
            & 0.00382  \ci{0.00421} & 0.04658  \ci{0.05770}
            & 0.04151  \ci{0.01943} \\
        \midrule
        \ours (ours)
            & 0.07661  \ci{0.26234} & 0.18295  \ci{0.23268}
            & 0.00152  \ci{0.00088} & 0.01794  \ci{0.01139}
            & 0.01770  \ci{0.00984} \\
        \ours SemEmb
            & 0.07880  \ci{0.22032} & 0.83306  \ci{0.74083}
            & 0.00348  \ci{0.00302} & 0.06406  \ci{0.09013}
            & 0.01321  \ci{0.00396} \\
        \midrule
        \ourrow\multicolumn{6}{l}{\textbf{(a) Ablation on Gaussian PDF}} \\
        \cdashline{1-6}\noalign{\vskip 0.6mm}
        \ours Gaussian
            & 0.08080  \ci{0.26208} & 0.13395  \ci{0.26072}
            & 0.02702  \ci{0.03854} & 0.32180  \ci{0.41047}
            & 0.05950  \ci{0.04229} \\
        \ours Gaussian PCGrad
            & 0.08141  \ci{0.30986} & 0.13132  \ci{0.25778}
            & 0.23146  \ci{0.52224} & 2.11588  \ci{4.26029}
            & 1.39698  \ci{3.32018} \\
        \midrule
        \ourrow\multicolumn{6}{l}{\textbf{(b) Ablation with Multi-Critic}} \\
        \cdashline{1-6}\noalign{\vskip 0.6mm}
        \ours MTCR
            & 0.10214  \ci{0.39199} & 0.30052  \ci{0.51518}
            & 0.00113  \ci{0.00090} & 0.01923  \ci{0.01473}
            & 0.01998  \ci{0.01062} \\
        \ours MTCR SemEmb
            & 0.07183  \ci{0.27774} & 0.41359  \ci{0.28789}
            & 0.00211  \ci{0.00138} & 0.03450  \ci{0.03229}
            & 0.01930  \ci{0.00855} \\
        \midrule
        \ourrow\multicolumn{6}{l}{\textbf{(c) Ablation with PCGrad}} \\
        \cdashline{1-6}\noalign{\vskip 0.6mm}
        \ours PCGrad s
            & 0.07860  \ci{0.29502} & 0.16608  \ci{0.24854}
            & 0.00355  \ci{0.00444} & 0.19859  \ci{0.20557}
            & 0.01771  \ci{0.00877} \\
        \ours PCGrad SemEmb
            & 0.07650  \ci{0.18997} & 0.60972  \ci{0.25001}
            & 0.00124  \ci{0.00077} & 0.01583  \ci{0.00990}
            & 0.02580  \ci{0.01580} \\
        \midrule
        \ourrow \multicolumn{6}{l}{\textbf{(d) Ablation with Multi-Critic and PCGrad}} \\
        \cdashline{1-6}\noalign{\vskip 0.6mm}
        \ours MTCR PCGrad
            & 0.06218  \ci{0.27415} & 0.20003  \ci{0.23535}
            & 0.00144  \ci{0.00143} & 0.01718  \ci{0.01882}
            & 0.01438  \ci{0.00515} \\
        \ours MTCR PCGrad SemEmb
            & 0.07261  \ci{0.26347} & 0.67473  \ci{0.35172}
            & 0.19082  \ci{0.09192} & 0.31052  \ci{0.09400}
            & 0.01660  \ci{0.00607} \\
        \bottomrule
        \end{tabular}
    }
    \label{tab:apx_abl_and_results_OoD_IntBall2}
\end{table}

\newpage

\begin{table*}[h]
    \centering
    \caption{\textsc{Reward functions used for learning GNC control tasks to train Int-Ball2.}}
    \renewcommand{\arraystretch}{1.05} 
    \resizebox{1\linewidth}{!}{
    \begin{tabular}{l l l l l}
    \toprule 
    Term & Expression & Weight & Scalar & Description \\
        
    \midrule 
    \noalign{\vskip -0.2mm}
      \ourrow \textbf{(a) Docking}  & $r_t^{\text{docking}} = r_t^{\text{pose}} + r_t^{v,\omega} + r_t^{\text{bnd}} + r^{\text{prog}}$ & & & \\ 
      
      \noalign{\vskip 0.4mm}\cdashline{1-5}\noalign{\vskip 0.8mm}
        Pose error & $\alpha_{\text{pose}} (\text{exp}(-d_p \backslash \lambda_p) \times \text{exp}(-d_h \backslash \lambda_h))$ & $\alpha_{\text{pose}}=2.0$ & $\lambda_p,\lambda_h = 1.0$ & Precise spatial convergence and alignment to target.\\

        Linear Velocity & $\alpha_{\text{v}} \text{clip}(v-v_{\text{min}}, 0, v_{\text{max}}-v_{\text{min}})$ & $\alpha_{\text{v}}=-0.08$ & $v_{\text{min}}=0.5, v_{\text{max}}=2.0$ & Velocity regulation and movement encouragement.\\

        Angular Velocity & $\alpha_{\omega} \text{clip}(\omega-\omega_{\text{min}}, 0, \omega_{\text{max}}-\omega_{\text{min}})$ & $\alpha_{\omega}=-0.08$ & $\omega_{\text{min}}=0.5, \omega_{\text{max}}=20.0$& Control of rotational stability.\\

        Boundary & $\alpha_{\text{bnd}} \text{exp}(-d_b \backslash \lambda_b)$ & $\alpha_{\text{bnd}}=-10.0$ & $\lambda_b=1.0$ & Safety constraint to stay within workspace.\\

        Progress & $\alpha_{\text{prog}}\Delta d_p (d_{\text{max}} - d_p)$ & $\alpha_{\text{prog}}=1.5$ & $ d_{\text{max}}=10.0 $ & Provides dense feedback for moving toward the goal.\\

    \midrule 
    \noalign{\vskip -0.2mm}
      \ourrow \textbf{(b) Track Velocities}  & $r_t^{\text{track}} = r_t^{\text{lin}} + r_t^{\text{ang}}$ & & & \\ 
      
      \noalign{\vskip 0.4mm}\cdashline{1-5}\noalign{\vskip 0.8mm}
        Surge error & $\beta_{\dot{x}} \text{exp}(-d_{\dot{x}} \backslash \lambda_{\dot{x}})$ & $\beta_{\dot{x}}=0.2$ & $\lambda_{\dot{x}}=1.0$ & Penalizes the absolute error between the target and current surge velocity.\\
        
        Sway error & $\beta_{\dot{y}} \text{exp}(-d_{\dot{y}} \backslash \lambda_{\dot{y}})$ & $\beta_{\dot{y}}=0.2$ & $\lambda_{\dot{y}}=1.0$ & Penalizes the absolute error between the target and current sway velocity.\\

        Heave error & $\beta_{\dot{z}} \text{exp}(-d_{\dot{z}} \backslash \lambda_{\dot{z}})$ & $\beta_{\dot{z}}=0.2$ & $\lambda_{\dot{z}}=1.0$ & Penalizes the absolute error between the target and current heave velocity.\\

        Roll error & $\beta_{\dot{\theta}} \text{exp}(-d_{\dot{\theta}} \backslash \lambda_{\dot{\theta}})$ & $\beta_{\dot{\theta}}=0.2$ & $\lambda_{\dot{\theta}}=1.0$ & Penalizes the absolute error between the target and current heading rotation rate.\\
        
        Pitch error & $\beta_{\dot{\phi}} \text{exp}(-d_{\dot{\phi}} \backslash \lambda_{\dot{y}})$ & $\beta_{\dot{\phi}}=0.2$ & $\lambda_{\dot{\phi}}=1.0$ & Penalizes the absolute error between the target and current pitch rotation rate.\\

        Yaw error & $\beta_{\dot{\psi}} \text{exp}(-d_{\dot{\psi}} \backslash \lambda_{\dot{\psi}})$ & $\beta_{\dot{\psi}}=0.2$ & $\lambda_{\dot{\psi}}=1.0$ & Penalizes the absolute error between the target and current roll rotation rate.\\

    \midrule 
    \noalign{\vskip -0.2mm}
      \ourrow \textbf{(c) Inspection}  & $r_t^{\text{insp}} = r_t^{\text{pose}} + r_t^{\text{prog}} + r_t^{v,\omega} + r_t^{\text{bnd}} + r_t^{\text{seq}}$ & & \\ 
      
      \noalign{\vskip 0.4mm}\cdashline{1-5}\noalign{\vskip 0.8mm}
        Pose error & $\gamma_{\text{pose}} (\text{exp}(-d_p \backslash \lambda_p) \times \text{exp}(-d_h \backslash \lambda_h))$ & $\gamma_{\text{pose}}=0.3$ & $\lambda_p, \lambda_h = 1.0$ & Precise spatial convergence and alignment to target.\\

        Linear Velocity & $\gamma_{\text{v}} \text{clip}(v-v_{\text{min}}, 0, v_{\text{max}}-v_{\text{min}})$ & $\gamma_{\text{v}}=-0.005$ & $v_{\text{min}}=0.5, v_{\text{max}}=2.0$ & Velocity regulation and movement encouragement.\\

        Angular Velocity & $\gamma_{\omega} \text{clip}(\omega-\omega_{\text{min}}, 0, \omega_{\text{max}}-\omega_{\text{min}})$ & $\gamma_{\omega}=-0.005$ & $\omega_{\text{min}}=0.5, \omega_{\text{max}}=20.0$ & Control of rotational stability.\\

        Boundary & $\gamma_{\text{bnd}} \text{exp}(-d_b \backslash \lambda_b)$ & $\gamma_{\text{bnd}}=-10.0$ & $\lambda_b=1.0$& Safety constraint to stay within workspace.\\

        Progress & $\gamma_{\text{prog}}\Delta d_p (d_{\text{max}} - d_p)$ & $\gamma_{\text{prog}}=3.0$ & $d_{\text{max}}=30.0$ & Provides dense feedback for moving toward the goal.\\

        Reach Goal & $\mathbf{1}(d_p < p)\times\mathbf{1}(d_{\theta} < o)$ & $\gamma_{\text{seq}}=10.0$ & $p=0.1, o=\frac{\pi}{20}$ & Bonus for every goal reached. \\

    \midrule 
    \noalign{\vskip -0.2mm}
      \ourrow \textbf{(d) Navigation with Obstacles}  & $r_t^{\text{nav}} = r_t^{\text{position}} + r_t^{v,\omega} + r_t^{\text{bnd}} + r_t^{\text{coll}}$ & & & \\ 

      \noalign{\vskip 0.4mm}\cdashline{1-5}\noalign{\vskip 0.8mm}
      Position error & $\zeta_{\text{position}} \text{exp}(-d_p \backslash \lambda_p)$ & $\zeta_{\text{position}}=1.0$ & $\lambda_p=0.8$& Precise spatial convergence and alignment to target.\\

        Linear Velocity & $\zeta_{\text{v}} \text{clip}(v-v_{\text{min}}, 0, v_{\text{max}}-v_{\text{min}})$ & $\zeta_{\text{v}}=-0.05$ & $v_{\text{min}}=0.5, v_{\text{max}}=2.0$ & Velocity regulation and movement encouragement.\\

        Angular Velocity & $\zeta_{\omega} \text{clip}(\omega-\omega_{\text{min}}, 0, \omega_{\text{max}}-\omega_{\text{min}})$ & $\zeta_{\omega}=-0.05$ & $\omega_{\text{min}}=0.5, \omega_{\text{max}}=20.0$ & Control of rotational stability.\\

        Boundary & $\zeta_{\text{bnd}} \text{exp}(-d_b \backslash \lambda_b)$ & $\zeta_{\text{bnd}}=-10.0$ & $ \lambda_b=1.0 $& Safety constraint to stay within workspace.\\

        Collision & $\mathbf{1}_{\text{coll}}$ & $\zeta_{\text{bnd}}=-50.0$ & & Penalty for colliding with an obstacle. \\

    \bottomrule
    \end{tabular}}
    \vspace{-0.1in}
    \label{table:reward_functions} 
\end{table*}

\begin{table*}[h!]
    \centering
    \setlength{\tabcolsep}{7pt}
    \renewcommand{\arraystretch}{1.2}
    \caption{\textsc{Performance of Int-Ball2 in simulation of all four tasks and five ablation studies of \ours.} \textbf{a)} Adds either the Task ID or the SemEmb ID into the observation space. \textbf{b)} Compares the sampling from \ours (beta distribution) vs the traditional sampling (Gaussian). \textbf{c)} Addition of the Multi-critic architecture into \ours. \textbf{d)} Addition of PCGrad into \ours. \textbf{e)} Addition of Multi-critic and PCGrad into \ours.}
    \resizebox{\textwidth}{!}{
        \begin{tabular}{l *{2}{c}  *{2}{c} *{2}{c}  *{1}{c}}
        \toprule
        \multirow{3}{*}{\textbf{Methods}}
        & \multicolumn{2}{c}{\textit{Docking}} 
        & \multicolumn{2}{c}{\textit{Track Velocities}} 
        & \multicolumn{2}{c}{\textit{Inspection}} 
        & \multicolumn{1}{c}{\textit{Navigation with Obstacles}} \\
        & $e_p [m] \downarrow$ & $e_o [rad] \downarrow$
        & $e_v [m\,s^{-1}] \downarrow$ & $e_{\omega} [rad\,s^{-1}] \downarrow$
        & $e_o [rad] \downarrow$ & $SR [\%] \uparrow$
        & $e_p [m] \downarrow$\\
        \midrule
        Single-Task RL  
            & 0.017 \ci{0.00773} & 0.041 \ci{0.02171}
            & 0.106 \ci{0.05330} & 0.178 \ci{0.06921}
            & 0.622 \ci{0.15115} & 85.46
            & 0.056 \ci{0.02538}\\
        \midrule
        Multi-Task RL 
            & 0.03684  \ci{0.02008} & 0.12672  \ci{0.06725}
            & 0.23349  \ci{0.09422} & 0.38662  \ci{0.03252}
            & 0.66191  \ci{0.15632} & 62.50
            & 0.11532  \ci{0.07620} \\
        Multi-Task RL TaskID
            & 0.01863  \ci{0.00795} & 0.03710  \ci{0.01281}
            & 0.18338  \ci{0.10423} & 0.19668  \ci{0.05987}
            & 0.57849  \ci{0.16597} & 83.750
            & 0.04357  \ci{0.02387} \\
        Multi-Task RL SemEmb
            & 0.02533  \ci{0.00863} & 0.03484  \ci{0.01461}
            & 0.18013  \ci{0.06820} & 0.20903  \ci{0.05861}
            & 0.62656  \ci{0.15893} & 87.5
            & 0.02533  \ci{0.00863} \\
        \midrule
        PCGrad  
            & 0.03230  \ci{0.01351} & 0.13668  \ci{0.06322}
            & 0.28194  \ci{0.10991} & 0.40364  \ci{0.03174}
            & 0.40000  \ci{0.49144} & 60.35
            & 0.11328  \ci{0.09077} \\
        PCGrad HotID  
            & 0.02256  \ci{0.01192} & 0.03703  \ci{0.01720}
            & 0.17720  \ci{0.07485} & 0.21131  \ci{0.06390}
            & 0.60587  \ci{0.17007} & 39.15
            & 0.03510  \ci{0.02124} \\
        PCGrad SemID  
            & 0.01923  \ci{0.00407} & 0.03398  \ci{0.02254}
            & 0.17901  \ci{0.07346} & 0.20544  \ci{0.06196}
            & 0.63376  \ci{0.18376} & 40.58
            & 0.04199  \ci{0.02098} \\
        \midrule
        Multi-Critic  
            & 0.02649  \ci{0.00477} & 0.04699  \ci{0.02683}
            & 0.18776  \ci{0.08791} & 0.24678  \ci{0.06491}
            & 0.58554  \ci{0.13618} & 85.00
            & 0.06019  \ci{0.03419} \\
        Multi-Critic TaskID  
            & 0.02094  \ci{0.00612} & 0.02397  \ci{0.01933}
            & 0.15095  \ci{0.07522} & 0.20349  \ci{0.07368}
            & 0.62124  \ci{0.17097} & 83.75
            & 0.09683  \ci{0.76961} \\
        Multi-Critic SemID  
            & 0.02324  \ci{0.01048} & 0.04599  \ci{0.01719}
            & 0.19854  \ci{0.09521} & 0.20509  \ci{0.06566}
            & 0.59556  \ci{0.17626} & 75.00
            & 0.04941  \ci{0.02244} \\
        \midrule
        \ours (ours)
            & 0.00566  \ci{0.00258} & 0.01154  \ci{0.00711}
            & 0.08610  \ci{0.04103} & 0.15713  \ci{0.05246}
            & 0.59716  \ci{0.15920} &  91.25
            & 0.02133  \ci{0.00891} \\
        \ourrow\multicolumn{8}{l}{\textbf{(a) Ablation \ours with TaskID and SemEmbID in the observation}} \\ 
        \cdashline{1-8}\noalign{\vskip 0.6mm}
        \ours (ours) Task ID
            & 0.00858  \ci{0.00638} & 0.01245  \ci{0.00581}
            & 0.10305  \ci{0.05323} & 0.14446  \ci{0.04740} 
            & 0.56139  \ci{0.14915} & 86.87
            & 0.08845  \ci{0.77098} \\
        \ours (ours) SemEmb ID
            & 0.00598  \ci{0.00230} & 0.00879  \ci{0.00476}
            & 0.09957  \ci{0.04856} & 0.14955  \ci{0.04903}
            & 0.56743  \ci{0.15128} & 88.12
            & 0.02626  \ci{0.01071} \\
        \midrule
        \ourrow\multicolumn{8}{l}{\textbf{(b) Ablation \ours with Gaussian PDF sampling}} \\ 
        \cdashline{1-8}\noalign{\vskip 0.6mm}
        \ours Gaussian
            & 0.01832  \ci{0.00889} & 0.03106  \ci{0.01685} 
            & 0.16635  \ci{0.06848} & 0.20587  \ci{0.06597}
            & 0.56204  \ci{0.16003} & 61.87
            & 0.11917  \ci{0.77562} \\
        \ours Gaussian PCGrad
            & 0.02434  \ci{0.00484} & 0.03718  \ci{0.01269}
            & 0.15313  \ci{0.05950} & 0.19770  \ci{0.06198}
            & 0.58275  \ci{0.14352} & 90.00
            & 0.23104  \ci{1.34745} \\
        \midrule
        \ourrow\multicolumn{8}{l}{\textbf{(c) Ablation \ours with Multi-Critic}} \\ 
        \cdashline{1-8}\noalign{\vskip 0.6mm}
        \ours MTCR   
            & 0.00706  \ci{0.00441} & 0.00901  \ci{0.00307}
            & 0.08193  \ci{0.05263} & 0.16291  \ci{0.04938}
            & 0.55893  \ci{0.14905} & 89.37
            & 0.08963  \ci{0.76815} \\
        \ours MTCR taskID   
            & 0.00647  \ci{0.00270} & 0.00912  \ci{0.00432}
            & 0.08143  \ci{0.05269} & 0.16279  \ci{0.05153}
            & 0.56170  \ci{0.13074} & 81.12
            & 0.01864  \ci{0.00742} \\
        \ours MTCR SemEmb  
            & 0.00938  \ci{0.00355} & 0.00940  \ci{0.00283}
            & 0.08652  \ci{0.04381} & 0.16601  \ci{0.04885}
            & 0.57676  \ci{0.15115} & 93.75
            & 0.01936  \ci{0.00854} \\
        \midrule
        \ourrow\multicolumn{8}{l}{\textbf{(d) Ablation \ours with PCGrad}} \\ 
        \cdashline{1-8}\noalign{\vskip 0.6mm}
        \ours PCGrad  
            & 0.01750  \ci{0.01220} & 0.00711  \ci{0.00387}
            & 0.09461  \ci{0.04953} & 0.21457  \ci{0.05988} 
            & 0.57654  \ci{0.14294} & 50.00
            & 0.08443  \ci{0.78843} \\
        \ours PCGrad TaskID  
            & 0.01301  \ci{0.00306} & 0.01445  \ci{0.00807}
            & 0.13753  \ci{0.07021} & 0.18106  \ci{0.05644}
            & 0.50334  \ci{0.21368} & 34.37
            & 0.09373  \ci{0.75429} \\
        \ours PCGrad SemEmb  
            & 0.02822  \ci{0.03512} & 0.02532  \ci{0.01529}
            & 0.11601  \ci{0.06506} & 0.16495  \ci{0.05245}
            & 0.56283  \ci{0.19814} & 82.50
            & 0.03287  \ci{0.02962} \\
        \midrule
        \ourrow\multicolumn{8}{l}{\textbf{(e) Ablation \ours with Multi-Critic and PCGrad}} \\ 
        \cdashline{1-8}\noalign{\vskip 0.6mm}
        \ours MTCR PCGrad  
            & 0.00284  \ci{0.00057} & 0.00678  \ci{0.00283}
            & 0.10765  \ci{0.05900} & 0.16990  \ci{0.05155}
            & 0.56633  \ci{0.14217} & 96.87
            & 0.01904  \ci{0.01073} \\
        \ours MTCR PCGrad TaskID  
            & 0.00211  \ci{0.00089} & 0.00832  \ci{0.00328}
            & 0.09460  \ci{0.05138} & 0.15136  \ci{0.04482}
            & 0.59240  \ci{0.15427} & 91.87
            & 0.01477  \ci{0.00777} \\
        \ours MTCR PCGrad SemEmb  
            & 0.00320  \ci{0.00170} & 0.00666  \ci{0.00354} 
            & 0.09738  \ci{0.04720} & 0.16592  \ci{0.05349}
            & 0.58588  \ci{0.16597} & 91.25
            & 0.01537  \ci{0.00686} \\
        \bottomrule
        \end{tabular}
    }
    \label{tab:apx_abl_and_results_IntBall2}
    \vspace{-5em}
\end{table*}

\end{document}